%% file: arxiv.tex
\documentclass[10pt,twocolumn,letterpaper]{article}
\def\withsupplement{1}
\def\xiapreprint{1}

\ifdefined\withsupplement\else
  \documentclass[10pt,twocolumn,letterpaper]{article}
\fi

\ifdefined\xiapreprint
  \usepackage[pagenumbers]{cvpr}
  \usepackage{balance}
\else
  \usepackage[review]{cvpr}
\fi
\input{preamble}

\ifdefined\withsupplement
  \usepackage{placeins}
\fi

\definecolor{cvprblue}{rgb}{0.21,0.49,0.74}
\ifdefined\xiapreprint
  \usepackage[breaklinks,colorlinks,allcolors=cvprblue,hypertexnames=false]{hyperref}
\else\ifdefined\overleaffast
  \setkeys{Gin}{draft}
  \usepackage[breaklinks,colorlinks,allcolors=cvprblue]{hyperref}
\else
  \usepackage[pagebackref,breaklinks,colorlinks,allcolors=cvprblue]{hyperref}
\fi\fi
\usepackage{algorithm}
\usepackage{algpseudocode}
\usepackage{xcolor}

\definecolor{XIAAlgGreen}{RGB}{24,122,70}

\algrenewcommand\algorithmiccomment[1]{%
    \hfill\textcolor{XIAAlgGreen}{$\triangleright$~#1}%
}

\def\confName{CVPR}
\def\confYear{2027}
\newcommand{\suppsection}[1]{%
  \ifdefined\withsupplement Supplement~\ref{#1}\else the supplementary material\fi}

\title{Beyond Expert Replay: Cross-Instruction Anchoring\\
for Continual Vision-Language-Action Learning}

\ifdefined\xiapreprint
  \input{arxiv_authors}
\else
  \author{Anonymous \confName~Submission}
\fi

\begin{document}
\ifdefined\xiapreprint
  \raggedbottom
\fi
\maketitle

\ifdefined\withsupplement
  \input{sec/paper_body}
\fi

\ifdefined\withsupplement
  \ifdefined\xiapreprint\else
    \clearpage
  \fi
\else
  \clearpage
  \onecolumn
  \twocolumn
\fi

\newcommand{\paperbibliography}{%
  {\small\bibliographystyle{ieeenat_fullname}\bibliography{main}}}
\ifdefined\xiapreprint\else
  \paperbibliography
\fi

\ifdefined\withsupplement
  \clearpage
  \appendix
  \def\continuoussupplement{1}
  \setcounter{figure}{0}
  \setcounter{table}{0}
  \renewcommand{\thefigure}{S\arabic{figure}}
  \renewcommand{\thetable}{S\arabic{table}}
  \renewcommand{\floatpagefraction}{0.6}
  \renewcommand{\dblfloatpagefraction}{0.6}
  \input{sec/S_supplement_compact}

\fi

\ifdefined\xiapreprint
  \onecolumn
  \twocolumn
  \balance
  {\interlinepenalty=10000 \paperbibliography}
\fi

\end{document}

%% file: preamble.tex
\usepackage[T1]{fontenc}
\usepackage{amsmath}
\usepackage{amssymb}
\usepackage{booktabs}
\usepackage{array}
\usepackage{tabularx}
\usepackage{multirow}
\usepackage{colortbl}
\usepackage{xspace}
\usepackage{microtype}
\usepackage{graphicx}

\definecolor{ResultBlue}{RGB}{43,101,154}
\definecolor{ResultBlueLight}{RGB}{231,240,248}
\definecolor{ResultGray}{RGB}{244,245,247}
\definecolor{ResultGrayDark}{RGB}{86,91,99}

\newcommand{\meanstd}[2]{#1\,\ensuremath{\pm}\,#2}

\newcolumntype{Y}{>{\centering\arraybackslash}X}
\newcolumntype{L}[1]{>{\raggedright\arraybackslash}p{#1}}
\newcolumntype{C}[1]{>{\centering\arraybackslash}p{#1}}

\renewcommand{\arraystretch}{1.10}

%% file: arxiv_authors.tex
\author{Ziyang Chen\textsuperscript{1,}\thanks{Equal contribution.},
Dehua Chen\textsuperscript{1,}\footnotemark[1],
Shaoguang Wang\textsuperscript{2},
Xu Yan\textsuperscript{1},
Hao Huang\textsuperscript{3},
Tao Chen\textsuperscript{1,}\thanks{Corresponding author.}\\[3pt]
\textsuperscript{1}Fudan University\quad
\textsuperscript{2}The Hong Kong University of Science and Technology (Guangzhou)\\[2pt]
\textsuperscript{3}New York University Abu Dhabi}

%% file: sec/paper_body.tex
\input{sec/0_abstract}
\input{sec/1_intro}
\input{sec/2_related_work}
\input{sec/limited_replay_analysis}
\input{sec/3_method}
\input{sec/4_experiments}
\input{sec/4_results}
\input{sec/5_conclusion}

%% file: sec/0_abstract.tex
\begin{abstract}
Vision-language-action (VLA) models need to continually acquire
new manipulation skills while retaining the ability to follow
previously learned instructions.
Experience replay (ER) can mitigate catastrophic forgetting, but retaining only
a small number of historical expert frames per task leaves many
old-instruction responses without direct supervision.
Under limited replay, we observe a structured failure mode in
which an earlier instruction can trigger a subsequently learned,
competing goal.
To address this gap, we propose Cross-Instruction
Anchoring (XIA), which supplements expert replay with
instruction-conditioned response preservation.
At each task boundary, XIA freezes the previous policy as a
teacher and matches its responses to old instructions paired
with current observations.
This extends supervision beyond stored historical examples
without requiring additional expert replay data.
Experiments across multiple VLA architectures and three
LIBERO suites demonstrate improved success rates and reduced
forgetting compared with ER under identical historical
expert-data budgets.
Additional RoboCasa and real-world experiments further support the effectiveness of XIA.
\end{abstract}

%% file: sec/1_intro.tex
\section{Introduction}
\label{sec:intro}

Vision-language-action (VLA) models unify visual perception,
language understanding, and action generation within a single robot
policy~\citep{brohan2023rt2,kim2024openvla,black2024pi0}.
By conditioning robot behavior on natural-language instructions,
they offer a flexible framework for diverse manipulation tasks.
As task requirements evolve after deployment, these policies must
continually acquire new skills while retaining the ability to follow
previously learned instructions.
\begin{figure}[!t]
\centering
\includegraphics[width=\columnwidth]{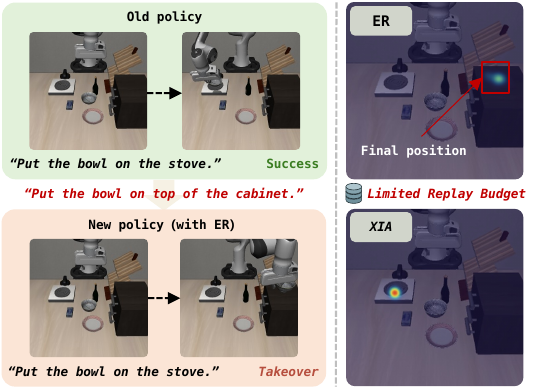}
\caption{\textbf{Instruction takeover under limited historical expert replay.}
\textbf{Left:} With a small historical expert replay buffer,
ER can redirect behavior toward a subsequently learned goal:
after learning to place the bowl on the cabinet, the policy
may pursue this goal under the earlier stove-placement instruction.
\textbf{Right:} Heatmaps of final bowl positions over 50
closed-loop trials per checkpoint, all evaluated under the
original stove-placement instruction.
Under the same replay budget, ER outcomes concentrate near
the cabinet, whereas XIA outcomes remain near the intended stove.}
\label{fig:instruction-takeover}
\end{figure}

One approach to continual VLA learning introduces task-adaptive
modules and routing mechanisms to mitigate forgetting, at the
cost of additional model capacity~\citep{romer2026clare}.
Experience replay (ER) offers a simpler alternative that
maintains a shared policy by interleaving new demonstrations
with stored historical expert samples.
Recent studies show that pretrained VLAs can retain previously
learned skills with simple ER~\citep{liu2026pretrained},
while selecting informative historical examples can further
improve replay effectiveness~\citep{du2026anchors}.
However, when only a small number of expert frames can be
retained per task, replay provides limited coverage of historical
observation--instruction pairs, and substantial forgetting
can still occur.
This raises a central question:
\emph{Can a shared VLA preserve its ability to follow
previously learned instructions under a tiny per-task
expert replay budget?}

\begin{figure}[!t]
\centering
\includegraphics[width=\columnwidth]{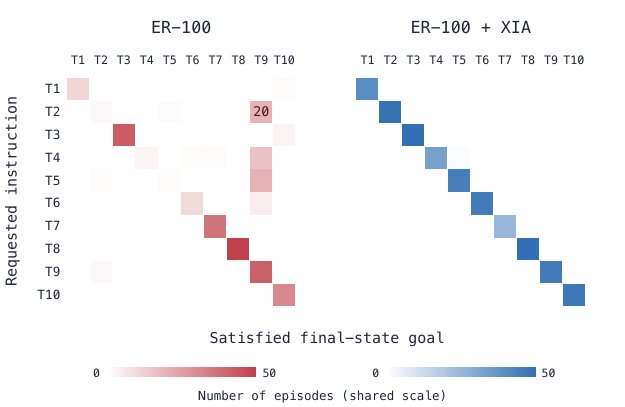}
\caption{\textbf{Instruction-conditioned final-goal outcomes.}
We evaluate OpenVLA-OFT after sequential training on ten
LIBERO-Goal tasks (50 episodes per instruction).
Rows indicate requested instructions; columns indicate goals
satisfied in the final simulator state.
Color intensity shows episode counts on a shared 0--50 scale;
episodes satisfying no goal are omitted.
ER-100 often reaches unintended goals (e.g., T9 in
20 episodes under T2), whereas ER-100 + XIA
concentrates outcomes along the intended diagonal.}
\label{fig:oft-goal-confusion}
\end{figure}

Under limited replay, forgetting can manifest as a systematic
shift in behavior rather than simply failing to complete an
earlier task. When prompted with a previously learned instruction,
a policy may instead pursue a competing goal learned later
(Figure~\ref{fig:instruction-takeover}), a phenomenon we term
\emph{instruction takeover}. Our analysis in Section~\ref{sec:replay-analysis} identifies
a fundamental limitation of fixed-buffer replay:
resampling or reweighting stored demonstrations cannot directly
supervise old-instruction responses on observations absent from
the replay buffer. Although generalization beyond stored examples
remains possible, these responses receive no explicit supervision
from ER. Consistent with this coverage limitation, we observe
$21.9\times$--$68.3\times$ larger response drift on unretained
observations than on buffered examples.
Final-goal analysis further reveals systematic shifts toward
later-learned goals under earlier instructions
(Figure~\ref{fig:oft-goal-confusion}).

To address this supervision gap, we introduce
\emph{Cross-Instruction Anchoring} (XIA), which supplements
expert replay with additional supervision on current observations
paired with previously learned instructions.
At each task boundary, we freeze the previous policy as a teacher
and query it using these cross-instruction pairs.
The student matches the teacher's instruction-conditioned
action responses while continuing to learn from current and
replayed expert demonstrations. Building on the response-preservation principle of
Learning without Forgetting~\citep{li2017lwf},
XIA explicitly anchors the previous policy's behavior
under old instructions on new observations.
Importantly, it does not use current expert actions as
supervision targets for substituted old instructions,
nor does it require storing additional historical expert frames.
The teacher is used only during training. We evaluate XIA across multiple VLA architectures and three
LIBERO suites, with additional experiments on RoboCasa and
a real-robot task sequence.
Controlled ablations further examine the effects of instruction
conditioning, query source, and historical replay budget. To summarize, our main contributions are as follows:

\begin{enumerate}

\item We identify and characterize \emph{instruction takeover},
a structured forgetting pattern in which earlier instructions
elicit behaviors directed toward subsequently learned goals,
and analyze the supervision-coverage limitation of
small-buffer experience replay.

\item We propose \emph{Cross-Instruction Anchoring} (XIA),
a conditional response-preservation objective that supplements
expert replay by anchoring the previous policy's responses
to old instructions on current observations, without storing
additional historical expert frames.

\item We evaluate XIA across multiple VLA architectures,
LIBERO task suites, RoboCasa, and real-world manipulation,
with controlled ablations of instruction conditioning,
query source, and historical replay budget.

\end{enumerate}

%% file: sec/2_related_work.tex
\section{Related Work}
\label{sec:related}

\paragraph{Language-Conditioned Manipulation and Instruction Grounding.}
Language-conditioned robot policies, such as BC-Z and RT-1, learn
shared manipulation behaviors from diverse language-annotated
demonstrations~\citep{jang2022bcz,brohan2022rt1}.
Recent vision-language-action (VLA) models generate actions through
discrete token prediction~\citep{brohan2023rt2,kim2024openvla},
diffusion~\citep{chi2023diffusionpolicy,octo2024},
flow matching~\citep{black2024pi0,physicalintelligence2025pi05,nvidia2025gr00t},
or direct regression~\citep{kim2025oft}.
Their capabilities are evaluated across benchmarks covering
long-horizon instruction following~\citep{mees2022calvin},
kitchen manipulation~\citep{nasiriany2024robocasa,nasiriany2026robocasa365},
dual-arm manipulation~\citep{mu2025robotwin},
and lifelong learning~\citep{liu2023libero}.

Beyond task completion, reliable language-conditioned control
requires policies to distinguish different goals within similar
visual scenes.
RoboGround uses grounding masks to identify target objects and
placement regions~\citep{huang2025roboground}, while CoT-VLA
generates visual subgoals to guide action prediction~\citep{zhao2025cotvla}.
LIBERO-Plus evaluates robustness to visual and linguistic
perturbations~\citep{fei2026liberoplus}, whereas LIBERO-CF
examines whether policies follow alternative instructions
under fixed scene layouts~\citep{fang2026liberocf}.
Related studies reveal that visual context can override
linguistic specifications~\citep{jiang2026groundinggap}.
More recently, \citet{raghavan2026grounded} investigate
instruction grounding throughout continual imitation learning.
Building on these observations, we examine how limited historical
replay affects the ability of a VLA to follow previously learned
instructions after sequential adaptation.

\paragraph{Continual Robot Learning.}
Continual robot learning aims to acquire new skills without
forgetting previously learned behaviors.
Existing approaches address this challenge through skill
reuse, model adaptation, and experience preservation.
LOTUS discovers reusable skills from demonstrations~\citep{wan2024lotus},
while Primitive Prompt Learning learns motion-aware prompts
shared across manipulation skills~\citep{yao2025ppl}.
CLARE selectively expands and routes task-specific
adapters~\citep{romer2026clare}, and Stellar VLA exploits
task--skill relationships for expert routing~\citep{wu2025stellar}.

For shared-policy adaptation, parameter
regularization~\citep{kirkpatrick2017overcoming} and experience
replay~\citep{rolnick2019experience} provide complementary
mechanisms for mitigating forgetting.
Recent studies show that simple replay can achieve strong
retention in pretrained VLAs~\citep{liu2026pretrained}
and support continual learning on physical robots~\citep{zhu2026realcontinual}.
Other methods improve the use of historical experience.
Memory Anchors prioritizes past demonstrations associated with
overlapping states and conflicting actions~\citep{du2026anchors},
while multimodal latent replay stores compact visual, linguistic,
and robot-state representations for subsequent
adaptation~\citep{yu2026latentreplay}.
Our work considers a complementary question:
how to strengthen experience replay when the amount of
retained historical expert data is limited.

\paragraph{Replay and Policy Distillation.}
Knowledge distillation mitigates forgetting by preserving
responses from a previous model.
Learning without Forgetting (LwF) distills old-task predictions
on new-task inputs~\citep{li2017lwf}, while Dark Experience Replay
matches predictions stored with historical
examples~\citep{buzzega2020dark}.
In robot learning, CIDER preserves a frozen actor's action
distributions on archived states~\citep{li2026cider},
and Info-VLA introduces teacher-guided constraints to maintain
cross-modal representations during continual
adaptation~\citep{zhao2026infovla}.
Building on response preservation, XIA queries the preceding
policy with current observations paired with previously
learned instructions and constrains the student's corresponding
action responses.
This provides additional instruction-conditioned supervision
alongside expert replay without requiring more historical
expert demonstrations.

%% file: sec/limited_replay_analysis.tex
\section{What Does Limited Replay Preserve?}
\label{sec:replay-analysis}
\label{sec:motivation}
\label{sec:rq1}

\paragraph{Old Instructions Can Trigger Later-learned Goals.}
We first examine a controlled two-task transition in LIBERO-Goal,
where QwenVLA learns to place a bowl on the stove and subsequently
on top of a cabinet.
Across two training seeds, the policy achieves 43/50 and 39/50
successes on the stove task before learning the cabinet task.
After adapting to the cabinet task with ER-100, however,
the original stove instruction triggers the competing cabinet
goal in 47/50 and 49/50 trials, respectively.
Doubling the number of replay samples from four to eight per
update, while keeping four current-task samples and the
historical buffer unchanged, does not eliminate this behavior.

Importantly, instruction takeover is not limited to this
two-task setting.
Figure~\ref{fig:oft-goal-confusion} shows final-goal outcomes
after a complete ten-task LIBERO-Goal sequence using OpenVLA-OFT.
Under ER-100, several earlier instructions lead to goals
learned later in the sequence, revealing similar takeover
behavior across different task pairs.
In contrast, ER-100 + XIA substantially reduces these
later-goal outcomes, with final goals more closely aligned
with the requested instructions.

\paragraph{Replay Frequency Does Not Expand Coverage.}
Let $P$ denote the training distribution over $(q,\ell,y)$
induced by current and replay examples, including data
augmentation.
Here, $q$ contains the observation $o$ and any additional
non-language inputs, $\ell$ denotes the instruction, and
$y$ is the supervision target.
Consider a set of observations $S$ that can occur during
old-task execution but satisfies
\[
P(o\in S,\ell=\ell_{\mathrm{old}})=0.
\]
Suppose two policies $f$ and $g$ agree everywhere except
possibly on these old-instruction--observation pairs.
Their expected training losses are then identical:

\begin{equation}
\mathbb E_P
\left[L(f(q,\ell),y)\right]
=
\mathbb E_P
\left[L(g(q,\ell),y)\right].
\label{eq:replay-support}
\end{equation}

\emph{Proof.}
Since $P$ assigns zero probability to the pairs on which
$f$ and $g$ may differ, their per-example losses agree
almost surely.
Resampling or reweighting examples within the same training
support cannot change this equality.
Even if observations in $S$ appear under a new instruction,
those samples do not directly supervise responses to
$\ell_{\mathrm{old}}$.
This characterizes a limitation of explicit replay
supervision without ruling out generalization to
unobserved pairs.

\paragraph{Response Drift beyond Stored Examples.}
We further examine how the QwenVLA policy changes before
and after learning the cabinet task.
Under the same old instruction, we fix the observation,
noisy action chunk, and flow time, and measure the
mean squared difference between the two policies'
predicted flow velocities.
This measures response drift rather than prediction
error against expert actions.

For each of the two seeds, we compare eight retained
frames with eight unretained frames drawn from the same
demonstration trajectories.
Both groups include observations from all four
trajectory-time quartiles.
The mean response drift on unretained frames is
$21.9\times$ and $68.3\times$ that on retained frames,
respectively. These measurements show that relatively stable responses
on stored replay examples can coexist with substantially
larger changes on unretained observations.
Together with the closed-loop evidence of instruction
takeover, they motivate extending old-instruction
response constraints beyond stored historical expert data.

%% file: sec/3_method.tex
\section{Cross-Instruction Anchoring}
\label{sec:method}
\label{sec:xia}

\input{figures/overview_20261112/figure}

Figure~\ref{fig:xia-overview} illustrates Cross-Instruction
Anchoring (XIA), which combines two complementary sources
of supervision for a shared student policy.
The expert replay path learns from current demonstrations
and a limited buffer of historical expert data.
The cross-instruction path preserves the preceding policy's
responses to previously learned instructions paired with
current observations.
At each task boundary, XIA freezes the preceding policy
as a teacher and jointly optimizes the student using
both objectives.

Following the response-preservation principle of Learning
without Forgetting~\citep{li2017lwf}, XIA introduces
additional constraints on instruction-conditioned action
predictions.
The current expert action is not used as the supervision
target for a substituted historical instruction.
Instead, the frozen teacher provides the corresponding
response target, extending supervision beyond stored
historical examples without requiring additional
expert demonstrations.

\subsection{Expert Replay Supervision}
\label{sec:er-objective}

Consider a sequence of $K$ manipulation tasks.
Let $D_k=\{(o_i,a_i,\ell_i)\}_{i=1}^{n_k}$
denote expert demonstration samples for task $k$,
where $o_i$, $a_i$, and $\ell_i$ represent the
observation, expert action, and language instruction,
respectively.
Let $\pi_\theta$ denote the trainable VLA policy
and $\mathcal B_{k-1}$ the replay buffer retaining
at most $M$ expert frames per previous task.

For task $k$, we sample a minibatch $\mathcal M_k$
from the current dataset $D_k$
and historical buffer $\mathcal B_{k-1}$.
When historical data are available, current and replay
samples are drawn in equal proportions.
Each sample retains its original instruction and
expert action.
The expert replay objective is

\begin{equation}
\mathcal L_{\mathrm{ER}}(\theta)
=
\frac{1}{|\mathcal M_k|}
\sum_{(o,a,\ell)\in\mathcal M_k}
\ell_{\mathrm{BC}}(\theta;o,a,\ell),
\label{eq:er}
\end{equation}

where $\ell_{\mathrm{BC}}$ is the backbone's native
behavior-cloning loss.
For flow-matching policies, it supervises the predicted
velocity field using targets constructed from expert
actions and sampled noise.
For OpenVLA-OFT, it applies L1 regression to normalized
action chunks.
For the first task, $\mathcal B_0=\varnothing$,
and training uses only current-task demonstrations.

Although replay constrains the policy on stored
observation--instruction pairs, many other observations
remain without direct supervision under historical
instructions.
XIA supplements expert replay with response constraints
on additional observation--instruction combinations.

\subsection{Cross-Instruction Response Preservation}
\label{sec:cross-response}

\paragraph{Cross-instruction Anchor Construction.}
At the start of task $k>1$, we freeze the preceding
policy $\theta_{k-1}$ as the boundary teacher $\theta^-$.
Let $\mathcal L_{<k}$ denote the set of previously
learned task instructions.
We uniformly sample up to $N$ frame indices
$\mathcal I_k$ from $D_k$ without replacement,
where $|\mathcal I_k|=\min(N,|D_k|)$.
The resulting anchor pool is

\begin{equation}
\begin{aligned}
\mathcal A_k
&= \{(o_i,a_i,\ell_i^-)\}_{i\in\mathcal I_k},\\
\ell_i^-
&\overset{\mathrm{i.i.d.}}{\sim}
\operatorname{Unif}(\mathcal L_{<k}),
\qquad k>1.
\end{aligned}
\label{eq:anchor-pool}
\end{equation}

Each sampled frame is paired with exactly one
historical task instruction, and this assignment
remains fixed throughout training on task $k$.
The current expert action $a_i$ is used only
to construct flow-matching queries, not as a
supervision target for the substituted instruction.
For the first task, we set $\mathcal A_1=\varnothing$
and $\mathcal L_{\mathrm{cross}}=0$.

\paragraph{Conditional Response Matching.}

Let $r_\theta(q,\ell)$ denote the policy's
action-head response to query $q$ conditioned
on instruction $\ell$.
The query and response depend on the
action-prediction architecture. For each anchor, the student and frozen teacher
receive the same query $q$ and historical instruction
$\ell^-$.
XIA minimizes their response discrepancy:

\begin{equation}
\mathcal L_{\mathrm{cross}}(\theta)
=
\mathbb E_{(q,\ell^-)\sim\rho_k}
\left[
d\left(
r_\theta(q,\ell^-),
\operatorname{sg}
\left[r_{\theta^-}(q,\ell^-)\right]
\right)
\right],
\label{eq:cross-response}
\end{equation}

where $\rho_k$ denotes the query distribution induced
by the anchor pool and architecture-specific query sampling.
The stop-gradient operator $\operatorname{sg}[\cdot]$
prevents gradients from propagating through the frozen teacher,
so only the student parameters are updated.
We define the response distance as

\begin{equation}
d(u,v)=
\begin{cases}
\displaystyle
\frac{1}{p}\|u-v\|_2^2,
& \text{flow-matching policies},
\\[6pt]
\displaystyle
\frac{1}{p}\|u-v\|_1,
& \text{OpenVLA-OFT},
\end{cases}
\label{eq:response-distance}
\end{equation}

where $p$ is the total number of scalar elements
in the predicted response tensor.
XIA directly matches action-head outputs rather than
intermediate hidden states.

\paragraph{Architecture-specific Response Queries.}

For flow-matching policies, XIA preserves the
predicted flow velocity:

\begin{equation}
r_\theta(q,\ell)
=
v_\theta(z_t,t,o,\ell),
\qquad
q=(z_t,t,o).
\label{eq:flow-response}
\end{equation}

The interpolated action $z_t$ is constructed from
the current expert action chunk $a$, Gaussian noise
$\epsilon\sim\mathcal N(0,I)$, and flow time $t$
sampled from the policy's native training distribution.
QwenVLA and $\pi_{0.5}$ use different interpolation
conventions:

\begin{equation}
\begin{aligned}
\text{QwenVLA:}\quad
z_t &= (1-t)\epsilon+ta,
& u_t &= a-\epsilon,
\\
\pi_{0.5}:\quad
z_t &= t\epsilon+(1-t)a,
& u_t &= \epsilon-a.
\end{aligned}
\label{eq:flow-conventions}
\end{equation}

Here $u_t$ is the velocity target used by the
standard behavior-cloning objective.
The cross-instruction loss instead uses the
frozen teacher's predicted velocity conditioned
on the historical instruction:

\begin{equation}
\operatorname{sg}
\left[
v_{\theta^-}(z_t,t,o,\ell^-)
\right].
\label{eq:teacher-flow-target}
\end{equation}

The current expert action therefore determines
the flow query $z_t$ but does not serve as the
supervision target for the substituted instruction.
Teacher and student are evaluated on identical
$(z_t,t,o,\ell^-)$ inputs.

For OpenVLA-OFT, no flow interpolation is required.
We set $q=o$, with $r_\theta(q,\ell)$ denoting
the predicted normalized action chunk.
The student matches the teacher's prediction
using the L1 distance in
Eq.~\ref{eq:response-distance}.

\paragraph{Joint Optimization.}

The student is trained by combining expert
replay with cross-instruction response preservation:

\begin{equation}
\boxed{
\mathcal L_{\mathrm{XIA}}(\theta)
=
\mathcal L_{\mathrm{ER}}(\theta)
+
\alpha\mathcal L_{\mathrm{cross}}(\theta)
}
\label{eq:xia}
\end{equation}

where $\alpha$ controls the strength of
cross-instruction anchoring.
Both losses are averaged independently before
being combined, so the number of anchor queries
does not alter the normalization of the
expert replay objective.

The two objectives provide complementary supervision:
ER fits expert actions on current and retained
historical examples, while the cross-instruction
term constrains the student's responses to old
instructions on additional observations.
These constraints preserve the preceding policy's
conditional predictions, but do not guarantee
that every teacher response is correct or that
every historical instruction is executable
from the queried observation.

\input{tables/xia_algorithm}

\begin{figure*}[!ht]
\centering
\includegraphics[width=\textwidth]{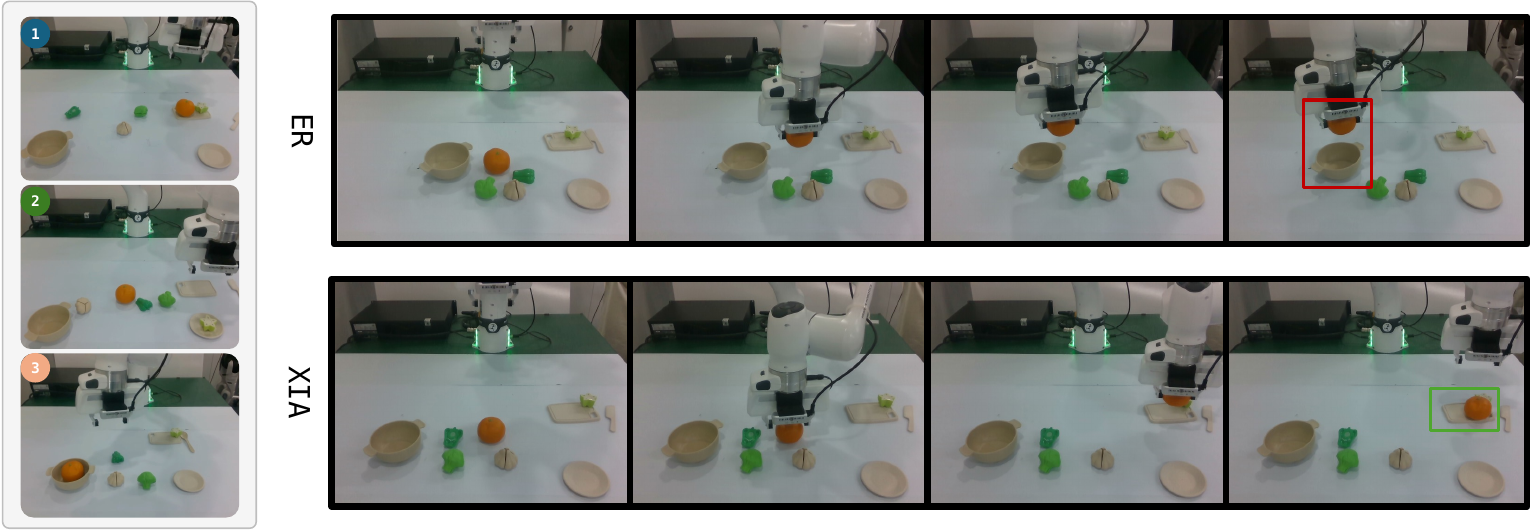}
\caption{\textbf{Real-robot task stream and qualitative comparison.}
\textbf{Left:} Final states of the three tasks in learning order.
\textbf{Right:} Example rollouts after learning all three tasks, evaluated
under the original T1 instruction ``pick up the orange and place it on the cutting board.''
ER-100 (top) moves the orange toward the bowl instead of the cutting board,
whereas XIA (bottom) correctly completes the original task.}
\label{fig:real-robot-tasks}
\end{figure*}
\subsection{Training Procedure and Inference}
\label{sec:xia-training}

Algorithm~\ref{alg:xia} summarizes the sequential
training procedure.
For the first task, no historical instructions
are available, and training reduces to behavior cloning.
For subsequent tasks,
$\operatorname{Anchor}_{N}^{<k}(D_k)$ constructs
the fixed anchor pool defined in Eq.~\ref{eq:anchor-pool}.
We set $\alpha=1$ and sample up to $N=1{,}000$
anchor frames per task.
At each optimizer update, four distinct anchors
are sampled, allowing reuse across updates.

For QwenVLA, we fix four noise--time tuples
per anchor and cache teacher velocities
for fixed, preprocessed images.
The student uses identical image and flow-query inputs.
For $\pi_{0.5}$ and OpenVLA-OFT, which employ
stochastic image augmentation, teacher responses
are computed online on the same augmented views
as the student.
For flow-matching policies, the cross-instruction
loss is averaged over anchors, noise--time tuples,
action steps, and coordinates.

After completing task $k$, we add up to $M$
unique expert frames from $D_k$ to the replay buffer.
At the next task boundary, the updated student
becomes the frozen teacher, while the anchor pool
and cached responses are discarded.
XIA retains the same historical expert data as ER
but incurs additional teacher computation and
temporary anchor storage during training.
At inference, only the student policy is used,
without the teacher, replay buffer, or anchor pool.
Implementation details are provided in
\suppsection{sec:xia-protocol}.

%% file: figures/overview_20261112/figure.tex
\begin{figure*}[!t]
\centering
\includegraphics[width=.70\textwidth]{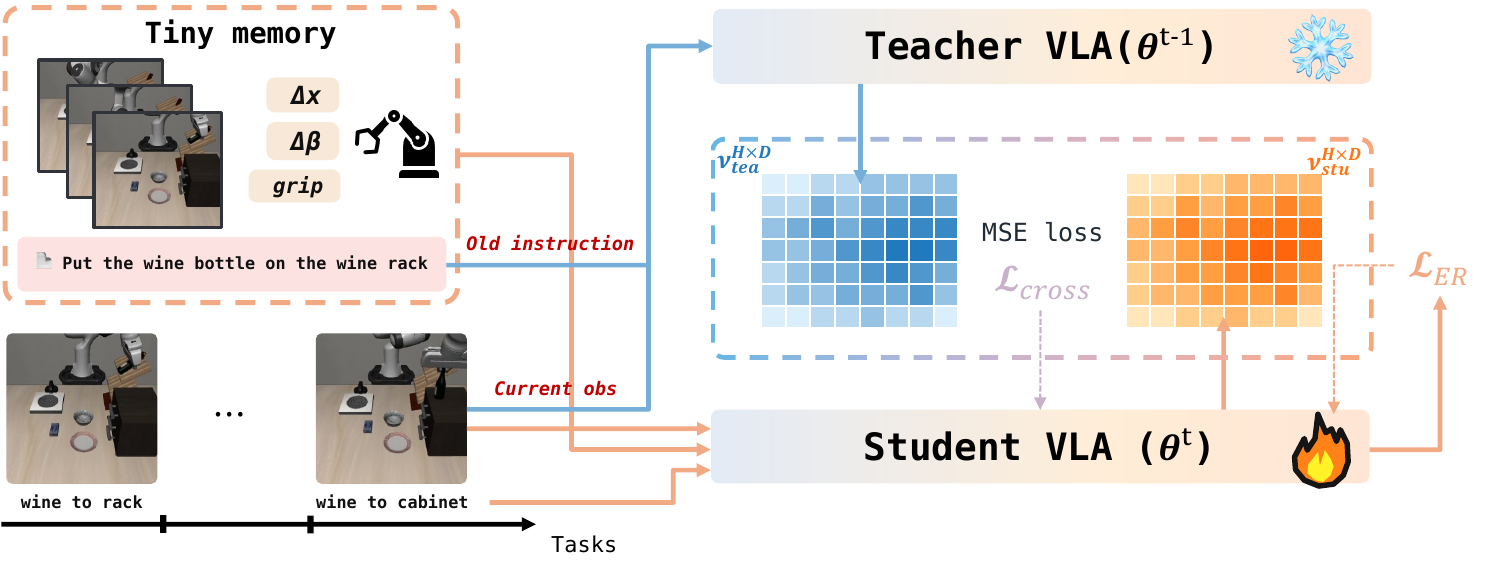}
\caption{\textbf{Overview of Cross-Instruction Anchoring (XIA).}
XIA combines two supervision paths under a limited historical
expert replay budget.
(1) \textbf{Expert Replay} ($\mathcal L_{\mathrm{ER}}$)
trains the student on current demonstrations and buffered
historical expert samples.
(2) \textbf{Cross-Instruction Anchoring} ($\mathcal L_{\mathrm{cross}}$)
pairs current observations with old instructions and aligns
student flow-velocity predictions with those of a frozen
task-boundary teacher.
Both objectives jointly update the student without additional
historical expert demonstrations.}
\label{fig:xia-overview}
\end{figure*}

%% file: tables/xia_algorithm.tex
\begin{algorithm}[h]
\caption{Cross-Instruction Anchoring (XIA)}
\label{alg:xia}

\footnotesize
\begin{algorithmic}[1]

\Require
Initial policy $\pi_{\theta_0}$;
task stream $\{D_k\}_{k=1}^{K}$

\Statex \hspace{\algorithmicindent}
$M$: expert replay frames per task;
$N$: anchor pool size

\Statex \hspace{\algorithmicindent}
$\alpha$: anchoring weight;
$\{U_k\}_{k=1}^{K}$: training steps

\State $\mathcal B_0 \gets \varnothing$

\For{$k=1,\ldots,K$}

    \State $\theta \gets \theta_{k-1}$

    \State $\mathcal A_k \gets \varnothing$

    \If{$k>1$}

        \State $\theta^- \gets
        \operatorname{Freeze}(\theta)$
        \Comment{Boundary teacher}

        \State $\mathcal A_k \gets
        \operatorname{Anchor}_{N}^{<k}(D_k)$
        \Comment{Old-text anchors}

    \EndIf

    \For{$u=1,\ldots,U_k$}

        \State $\mathcal M \sim
        \operatorname{Mix}(D_k,\mathcal B_{k-1})$

        \State $L_e \gets
        \mathcal L_{\mathrm{ER}}(\theta;\mathcal M)$

        \State $L_x \gets 0$

        \If{$\mathcal A_k \neq \varnothing$}

            \State $\mathcal Q \gets
            \operatorname{Batch}_4(\mathcal A_k)$

            \State $L_x \gets
            \mathcal L_{\mathrm{cross}}
            (\theta,\theta^-;\mathcal Q)$

        \EndIf

        \State $\theta \gets
        \operatorname{OptStep}
        (\theta,L_e+\alpha L_x)$
        \Comment{Joint update}

    \EndFor

    \State $\theta_k \gets \theta$

    \State $\mathcal S_k \gets
    \operatorname{Select}_M(D_k)$
    \Comment{Store expert frames}

    \State $\mathcal B_k \gets
    \mathcal B_{k-1}\cup\mathcal S_k$

\EndFor

\State \Return $\pi_{\theta_K}$
\Comment{Teacher-free inference}

\end{algorithmic}
\end{algorithm}

%% file: sec/4_experiments.tex
\section{Experiments}
\label{sec:experiments}
\label{sec:rq3}

\input{tables/goal10_main}
\subsection{Experimental Setup}
\label{sec:setup}

\paragraph{LIBERO Benchmark.}
Our primary evaluation uses ten-task continual learning sequences
from LIBERO-Goal, LIBERO-Object, and LIBERO-Spatial~\citep{liu2023libero}.
These suites emphasize variations in manipulation goals,
target objects, and spatial relationships, respectively.
We evaluate three VLA architectures:
QwenVLA, $\pi_{0.5}$, and OpenVLA-OFT.
QwenVLA combines Qwen2.5-VL-3B with a DiT-based flow-matching
action head, using rank-32 LoRA and 10,000 training updates
per task; it is initialized from a VLM rather than a
robot-pretrained GR00T checkpoint.
The flow-matching policy $\pi_{0.5}$ uses 7,500 updates per task,
while OpenVLA-OFT is initialized from OpenVLA-7B with a
continuous action regression head~\citep{kim2025oft}.
Each policy retains its native training objective and
optimizer configuration.

\paragraph{RoboCasa Benchmark.}
We further evaluate XIA on RoboCasa v0.2, a simulated kitchen
manipulation benchmark featuring interactive appliances
and articulated fixtures.
Drawing on the task selection in~\citep{hu2026simplerecipe},
we construct a five-task sequence comprising
(T1) sink-spout rotation, (T2) faucet activation,
(T3) microwave activation, (T4) microwave deactivation,
and (T5) faucet deactivation.
The sequence includes related manipulation tasks with
different objectives involving shared appliances.
We use GR00T N1.5-3B, sequentially fine-tuning its
action module while freezing the vision-language backbone.
Final checkpoints are evaluated over 50 rollouts per task
using one matched training seed.

\paragraph{Franka Real-Robot Experiments.}
We conduct real-world experiments using a Franka Research 3
(FR3) robot equipped with a gripper, an external RGB camera,
and a wrist-mounted RGB camera.
The three-task sequence consists of
(T1) placing an orange on a cutting board,
(T2) transferring sliced starfruit from the board to a plate,
and (T3) placing an orange in a bowl
(Figure~\ref{fig:real-robot-tasks}).
T1 and T3 share the manipulated object but require
different placement destinations.
We collect 184 language-annotated demonstration trajectories
across the three tasks.
Experiments use a DROID-initialized $\pi_{0.5}$ policy
with visual and proprioceptive observations, a 16-step
action horizon, and 10,000 training updates per task.

\paragraph{Baselines and Evaluation.}
We compare ER and ER + XIA under matched historical
expert-data budgets, retaining $M=100$ unique expert
frames per previous task with identical replay buffers.
XIA additionally distills frozen-teacher responses on
up to 1,000 current observations paired with old instructions. For LIBERO, we evaluate each task over 50 closed-loop
episodes. Final success rate (SR, \%) averages performance
across all $K=10$ tasks at the final checkpoint.
Let $R_{b,t}\in[0,1]$ denote task $t$ success after
training through task $b$.
Following LIBERO's task-wise formulation, we compute
negative backward transfer (NBT) as
\begin{equation}
\mathrm{NBT}=\frac{1}{K-1}\sum_{t=1}^{K-1}
\left(R_{t,t}-\frac{1}{K-t}\sum_{b=t+1}^{K}R_{b,t}\right).
\end{equation}
NBT averages signed performance drops over subsequent
task boundaries, allowing negative values for improvement.

On QwenVLA, we further compare ER + EWC with Fisher-weighted
regularization~\citep{kirkpatrick2017overcoming},
ER + SAM with sharpness-aware
optimization~\citep{foret2021sharpness},
DER++-style response distillation~\citep{buzzega2020dark},
and Info-VLA adapted from its official
implementation~\citep{zhao2026infovla}.
We report means and sample standard deviations over
three seeds, with details in
\suppsection{sec:seed-results}.

%% file: tables/goal10_main.tex
\begin{table*}[!ht]
\centering
\caption{\textbf{Continual learning with matched expert replay.}
Results on ten-task LIBERO-Goal, Object, and Spatial sequences
with 100 historical expert frames per task.
SR (\%) averages success rates across all ten tasks using the
final checkpoint, while Avg. SR averages SR across the three suites.
NBT measures the average performance drop from task acquisition
to subsequent learning boundaries (lower is better).
Suite-level results are reported as mean $\pm$ standard deviation
over three seeds.}

\label{tab:main-results}

\footnotesize
\setlength{\tabcolsep}{4pt}
\renewcommand{\arraystretch}{1.18}

\definecolor{XIALightBlue}{RGB}{232,243,250}

\begin{tabular}{@{}c|c||c|cc|cc|cc@{}}
\toprule

\multirow{2}{*}{\textbf{Policy}} &
\multirow{2}{*}{\textbf{Method}} &
\multirow{2}{*}{\textbf{Avg. SR} $\uparrow$} &
\multicolumn{2}{c|}{\textbf{LIBERO Goal}} &
\multicolumn{2}{c|}{\textbf{LIBERO Object}} &
\multicolumn{2}{c}{\textbf{LIBERO Spatial}}\\

& & &
\textbf{SR} $\uparrow$ & \textbf{NBT} $\downarrow$ &
\textbf{SR} $\uparrow$ & \textbf{NBT} $\downarrow$ &
\textbf{SR} $\uparrow$ & \textbf{NBT} $\downarrow$\\

\midrule

\multirow{2}{*}{QwenVLA}
& \cellcolor{ResultGray}ER
& \cellcolor{ResultGray}52.7
& \cellcolor{ResultGray}\meanstd{44.0}{3.6}
& \cellcolor{ResultGray}\meanstd{.45}{.06}
& \cellcolor{ResultGray}\meanstd{62.8}{3.0}
& \cellcolor{ResultGray}\meanstd{.33}{.02}
& \cellcolor{ResultGray}\meanstd{51.4}{2.4}
& \cellcolor{ResultGray}\meanstd{.37}{.04}\\

& \cellcolor{XIALightBlue}ER + XIA
& \cellcolor{XIALightBlue}\textbf{71.5}
& \cellcolor{XIALightBlue}\meanstd{\textbf{71.4}}{8.8}
& \cellcolor{XIALightBlue}\meanstd{\textbf{.16}}{.06}
& \cellcolor{XIALightBlue}\meanstd{\textbf{74.7}}{4.5}
& \cellcolor{XIALightBlue}\meanstd{\textbf{.18}}{.03}
& \cellcolor{XIALightBlue}\meanstd{\textbf{68.5}}{2.7}
& \cellcolor{XIALightBlue}\meanstd{\textbf{.22}}{.01}\\

\midrule

\multirow{2}{*}{$\pi_{0.5}$}
& \cellcolor{ResultGray}ER
& \cellcolor{ResultGray}82.5
& \cellcolor{ResultGray}\meanstd{74.6}{4.9}
& \cellcolor{ResultGray}\meanstd{.17}{.03}
& \cellcolor{ResultGray}\meanstd{88.8}{0.8}
& \cellcolor{ResultGray}\meanstd{.12}{.02}
& \cellcolor{ResultGray}\meanstd{84.0}{3.7}
& \cellcolor{ResultGray}\meanstd{.10}{.04}\\

& \cellcolor{XIALightBlue}ER + XIA
& \cellcolor{XIALightBlue}\textbf{89.0}
& \cellcolor{XIALightBlue}\meanstd{\textbf{80.0}}{8.2}
& \cellcolor{XIALightBlue}\meanstd{\textbf{.11}}{.06}
& \cellcolor{XIALightBlue}\meanstd{\textbf{91.3}}{1.0}
& \cellcolor{XIALightBlue}\meanstd{\textbf{.09}}{.04}
& \cellcolor{XIALightBlue}\meanstd{\textbf{95.8}}{0.5}
& \cellcolor{XIALightBlue}\meanstd{\textbf{.01}}{.02}\\

\midrule

\multirow{2}{*}{OpenVLA-OFT}
& \cellcolor{ResultGray}ER
& \cellcolor{ResultGray}72.5
& \cellcolor{ResultGray}\meanstd{50.3}{4.5}
& \cellcolor{ResultGray}\meanstd{.38}{.11}
& \cellcolor{ResultGray}\meanstd{84.9}{7.9}
& \cellcolor{ResultGray}\meanstd{.10}{.05}
& \cellcolor{ResultGray}\meanstd{82.4}{10.7}
& \cellcolor{ResultGray}\meanstd{.16}{.11}\\

& \cellcolor{XIALightBlue}ER + XIA
& \cellcolor{XIALightBlue}\textbf{92.6}
& \cellcolor{XIALightBlue}\meanstd{\textbf{88.0}}{3.5}
& \cellcolor{XIALightBlue}\meanstd{\textbf{.05}}{.07}
& \cellcolor{XIALightBlue}\meanstd{\textbf{95.5}}{3.5}
& \cellcolor{XIALightBlue}\meanstd{\textbf{.03}}{.03}
& \cellcolor{XIALightBlue}\meanstd{\textbf{94.2}}{2.5}
& \cellcolor{XIALightBlue}\meanstd{\textbf{.02}}{.03}\\

\bottomrule
\end{tabular}
\end{table*}

%% file: sec/4_results.tex
\subsection{Quantitative Results}
\label{sec:full-comparisons}

\paragraph{LIBERO.}
Table~\ref{tab:main-results} compares ER and ER + XIA
across three VLA policies and three LIBERO suites under
identical historical expert replay budgets.
XIA achieves higher final success rates and lower NBT
in all nine policy--suite combinations.
Averaged across suites, XIA improves final SR by
18.8, 6.5, and 20.1 percentage points for QwenVLA,
$\pi_{0.5}$, and OpenVLA-OFT, respectively.
The gains are particularly pronounced on LIBERO-Goal,
where SR increases from 44.0\% to 71.4\% for QwenVLA
and from 50.3\% to 88.0\% for OpenVLA-OFT.
These improvements are consistent with the
instruction takeover observed in our analysis:
limited historical replay may leave previous
instruction-conditioned behaviors insufficiently
constrained.
By preserving teacher responses to old instructions
on current observations, XIA provides complementary
supervision that improves retention under the same
expert replay budget.

\input{tables/goal_baselines_question}
We further compare our method with other continual learning approaches on QwenVLA. As shown in Table~\ref{tab:extended-baselines}, XIA achieves the highest mean final success rate across all three LIBERO suites, outperforming the evaluated regularization- and replay-based alternatives. Notably, ER + EWC emerges as the strongest competing baseline on average and outperforms the adapted Info-VLA method across all three suites. This finding suggests that recently proposed VLA-specific continual-learning methods do not necessarily outperform classical approaches when access to historical expert data is restricted. Overall, XIA provides consistent improvements under the same limited replay budget, highlighting the effectiveness of cross-instruction anchoring in preserving previously learned behaviors.

\begin{table}[h]
\centering
\caption{\textbf{Continual learning on RoboCasa.}
Final task success rates (\%) of GR00T N1.5-3B after
sequentially learning five tasks. Both methods use
100 historical expert frames per task.
Results are based on 
50 evaluation rollouts per task.}
\label{tab:groot-robocasa-final50}
\small
\setlength{\tabcolsep}{4pt}
\renewcommand{\arraystretch}{1.12}
\begin{tabular}{@{}lrrr@{}}
\toprule
\textbf{Task}
& \textbf{ER-100}
& \textbf{ER + XIA}
& $\boldsymbol{\Delta}$ \\
\midrule
T1: Turn sink spout
& 54.0 & \textbf{80.0} & +26.0 \\
T2: Faucet on
& 12.0 & \textbf{30.0} & +18.0 \\
T3: Microwave on
& 0.0 & \textbf{2.0} & +2.0 \\
T4: Microwave off
& \textbf{6.0} & 4.0 & -2.0 \\
T5: Faucet off
& \textbf{46.0} & 44.0 & -2.0 \\
\midrule
\textbf{Average success rate}
& 23.6 & \textbf{32.0} & +8.4 \\
\bottomrule
\end{tabular}
\end{table}

\paragraph{RoboCasa.}
Table~\ref{tab:groot-robocasa-final50} presents the RoboCasa results.
Under limited historical expert replay budget, XIA can help retain old task performance, improving
the final average success rate from 23.6\% to 32.0\%. These results provide additional evidence that XIA can improve retention across different VLA
architectures and manipulation benchmarks.

\paragraph{Real-World Experiments.}
\label{sec:real-robot}

\begin{table}[h]
\centering
\caption{\textbf{Real-robot evaluation.} Task success rates (\%) over 20 trials per learned task after each training stage. T1--T3 denote orange-to-board, starfruit-to-plate, and orange-to-bowl, respectively. XIA uses the same ER-100 replay budget. Final SR averages success across all three tasks after T3.}
\label{tab:real-robot-results}
\input{tables/real_robot_tally_20261116}
\end{table}

We also conduct a real-world continual learning experiment. As shown in Table~\ref{tab:real-robot-results}, XIA largely improves the final average success rate and mitigates forgetting. Qualitative analysis further reveals goal interference consistent with our simulation findings. At the final ER-100 checkpoint, the robot moves the orange toward the bowl in 12 of 20 T1 trials, despite being instructed to place it on the cutting board. These observations further support the robustness of XIA in preserving previously learned behaviors in a real-world scenario.

\subsection{Ablation Study}
\label{sec:conditioning}
\paragraph{Instruction and Query Source.}
\label{sec:query-limits}

\begin{figure}[h]
\centering
\includegraphics[width=\columnwidth]{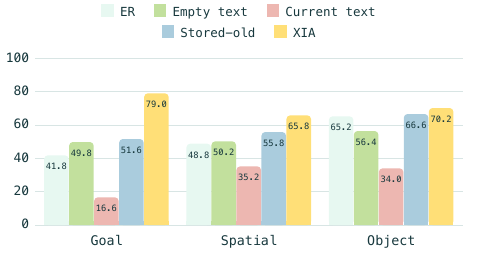}
\caption{\textbf{Ablations of instruction conditioning and query source.}
Empty-text and Current-text replace XIA's historical instruction with an empty prompt or the current-task instruction while keeping current-task observations. Stored-old holds the historical instruction but uses stored expert observations and their associated action-query endpoints. We report the final average success rate (\%) across ten tasks for QwenVLA on each LIBERO suite.}
\label{fig:full-sequence-ablation}
\end{figure}

We first examine instruction conditioning by replacing XIA's historical instructions with empty text or current-task instructions while keeping the current observations fixed. We then retain the historical instructions but construct teacher queries from stored expert demonstrations instead of current-task data (Stored-old). As shown in Figure~\ref{fig:full-sequence-ablation}, the full XIA achieves the highest final success rate across all three LIBERO suites. Notably, Stored-old consistently ranks second and numerically outperforms ER, although the improvement on Object is modest. In contrast, empty-text anchoring yields mixed results, whereas current-text anchoring performs worse than ER across all three suites. Since Stored-old also changes the action-query endpoints, its performance gap relative to XIA cannot be attributed exclusively to observation coverage. The results suggest that effective anchoring requires appropriately conditioned teacher responses, with additional benefits from querying beyond stored historical observations.

\paragraph{Sensitivity to Historical Expert-Frame Budget.}
\begin{figure}[h]
\centering
\includegraphics[width=\columnwidth]{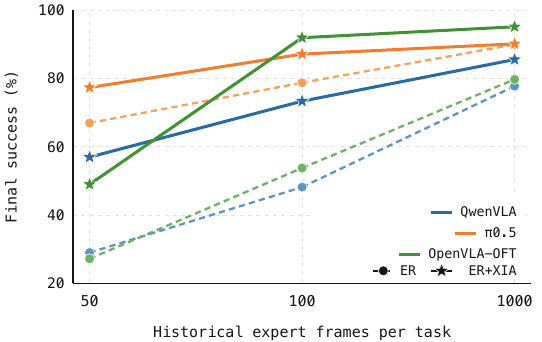}
\caption{\textbf{Historical expert-frame budget on LIBERO-Goal.}
Final success over 500 episodes per run. Colors denote policies; dashed
circles denote ER and solid stars ER + XIA.}
\label{fig:budget-50-500}
\end{figure}

Moreover, we investigate the effect of historical replay budget by varying the number of retained expert frames per task. As shown in Figure~\ref{fig:budget-50-500}, XIA consistently outperforms ER baseline across the evaluated budgets and policy architectures. Notably, the improvements are particularly pronounced under constrained replay budgets, where limited historical observations provide insufficient coverage of previously learned behaviors. This suggests that XIA offers complementary supervision through old-instruction-conditioned teacher responses, alleviating the reliance on extensive historical expert demonstrations.  Overall, these results demonstrate that XIA is especially effective in low-memory regimes.

%% file: tables/goal_baselines_question.tex
\begin{table}[h]
\centering
\caption{\textbf{Comparison with additional continual-learning baselines.} We present final success rates (SR, \%) of QwenVLA after learning whole ten task sequences in each LIBERO suite 
with 100 historical expert frames per task.}
\label{tab:extended-baselines}

\footnotesize
\setlength{\tabcolsep}{3pt}
\renewcommand{\arraystretch}{1.16}

\definecolor{XIALightBlue}{RGB}{232,243,250}

\begin{tabular}{@{}l|c|ccc@{}}
\toprule
\textbf{Method}
& \textbf{Avg.} $\uparrow$
& \textbf{Goal}
& \textbf{Spatial}
& \textbf{Object} \\
\midrule

\cellcolor{ResultGray}ER
& \cellcolor{ResultGray}52.7
& \cellcolor{ResultGray}\meanstd{44.0}{3.6}
& \cellcolor{ResultGray}\meanstd{51.4}{2.4}
& \cellcolor{ResultGray}\meanstd{62.8}{3.0} \\

ER + EWC
& \underline{65.9}
& \meanstd{\underline{61.7}}{5.3}
& \meanstd{61.4}{5.1}
& \meanstd{\underline{74.5}}{8.6} \\

ER + SAM
& 61.7
& \meanstd{50.5}{5.3}
& \meanstd{\underline{61.9}}{1.7}
& \meanstd{72.8}{5.9} \\

DER++-style
& 55.8
& \meanstd{44.2}{8.8}
& \meanstd{56.7}{3.5}
& \meanstd{66.4}{1.4} \\

Info-VLA$^\dagger$
& 53.8
& \meanstd{44.9}{8.9}
& \meanstd{50.9}{6.4}
& \meanstd{65.7}{2.7} \\

\addlinespace[2pt]

\rowcolor{XIALightBlue}
ER + XIA
& \textbf{71.5}
& \meanstd{\textbf{71.4}}{8.8}
& \meanstd{\textbf{68.5}}{2.7}
& \meanstd{\textbf{74.7}}{4.5} \\

\bottomrule
\end{tabular}

\vspace{3pt}
\parbox{\linewidth}{
\raggedright\scriptsize
$^\dagger$ Info-VLA results are obtained by adapting the authors' official implementation to our fixed-budget continual-learning setting.
}

\end{table}

%% file: tables/real_robot_tally_20261116.tex
\centering
\footnotesize
\setlength{\tabcolsep}{2pt}
\renewcommand{\arraystretch}{1.12}

\begingroup
\centering
\small
\setlength{\tabcolsep}{4pt}
\renewcommand{\arraystretch}{1.06}
\begin{tabular}{@{}lccc@{\hspace{1.1em}}ccc@{}}
  \toprule
  & \multicolumn{3}{c}{\textbf{ER-100}}
  & \multicolumn{3}{c}{\textbf{ER-100 + XIA}} \\
  \cmidrule(lr){2-4}\cmidrule(l){5-7}
  \textbf{Stage} & T1 & T2 & T3 & T1 & T2 & T3 \\
  \midrule
  After T1 & 90 & -- & --  & 90 & -- & -- \\
  After T2 & 75 & 95 & --  & \textbf{85} & \textbf{100} & -- \\
  After T3 & 20 & 60 & 100 & \textbf{95} & \textbf{80} & 100 \\
  \midrule
  Final SR (\%) $\uparrow$
    & \multicolumn{3}{c}{60.0} & \multicolumn{3}{c}{\textbf{91.7}} \\
  NBT $\downarrow$
    & \multicolumn{3}{c}{0.388} & \multicolumn{3}{c}{\textbf{0.100}} \\
  \bottomrule
\end{tabular}\par
\endgroup

%% file: sec/5_conclusion.tex
\section{Conclusion}
\label{sec:conclusion}

We present Cross-Instruction Anchoring (XIA) to improve
continual VLA learning under limited historical expert replay.
Our analysis reveals instruction takeover, where previously
learned instructions can trigger competing goals acquired later.
XIA preserves the preceding policy's responses to old
instructions on current observations, extending supervision
beyond stored expert demonstrations without requiring
additional historical expert data.
Experiments across multiple VLA architectures, LIBERO and
RoboCasa benchmarks, and real-world manipulation show
improved old-task retention under matched replay budgets.
These findings highlight that instruction-conditioned response
preservation as a useful complement to conventional
expert replay strategy.

%% file: sec/S_supplement_compact.tex
\ifdefined\continuoussupplement
  \twocolumn[{\centering\Large\bfseries Supplementary Material\par\medskip}]
\else
  \setcounter{page}{1}
  \maketitlesupplementary
\fi

\noindent This supplement presents the experimental protocol
(Sec.~\ref{sec:protocol-details}), additional results and controls
(Sec.~\ref{sec:seed-results}), and Franka demonstrations and policy
rollouts (Sec.~\ref{sec:franka-supplement}).

\input{sec/A_protocol}

\FloatBarrier
\section{Additional Results and Controls}
\label{sec:seed-results}

\subsection{Full-Sequence Retention}
\input{tables/sequence_forgetting}

Let $R_{b,t}$ be the success rate on task $t$ after training through
task $b$. Acquisition is $100K^{-1}\sum_{t=1}^{K} R_{t,t}$ in percent.
Normalized NBT divides each task's mean signed drop over later boundaries
by its acquisition rate $R_{t,t}$, then averages over tasks with later
boundaries and nonzero acquisition.

Figure~5 of the main paper uses one training seed per suite:
2024 for Goal and 42 for Object and Spatial. For $\pi_{0.5}$ Goal,
the paired final-success gains with XIA are $+8.4$, $+6.0$, and $+1.8$
percentage points for seeds 42, 1337, and 2024, respectively
(\meanstd{+5.4}{3.3} percentage points across seeds).

\subsection{Historical Replay Budget}
\begin{figure}[!htbp]
\centering
\includegraphics[width=.98\columnwidth]{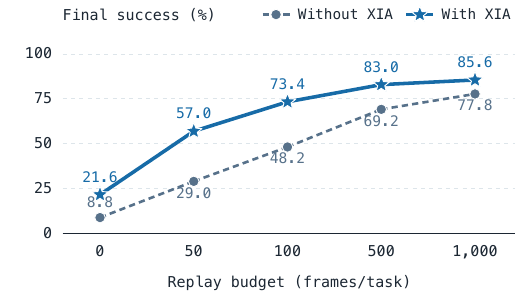}
\caption{\textbf{Final success across five historical replay budgets.}
QwenVLA on LIBERO-Goal, seed 42, 500 episodes per arm. The horizontal
axis shows discrete budget categories. With no replay, the baseline is sequential
fine-tuning. At nominal $M=1{,}000$, both arms retain 998 frames for one task.}
\label{fig:supp-goal-budget}
\end{figure}

Figure~\ref{fig:supp-goal-budget} extends the QwenVLA Goal comparison to
five replay budgets. At zero replay, XIA raises final success from
8.8\% to 21.6\% while using no old expert frames. At 500 frames per
task, it raises success from 69.2\% to 83.0\%. In this seed, XIA with
100 frames exceeds ER with 500 frames by 4.2 percentage points.

\subsection{Instruction Selection from Matched States}
\begin{figure}[!htbp]
\centering
\includegraphics[width=.98\columnwidth]{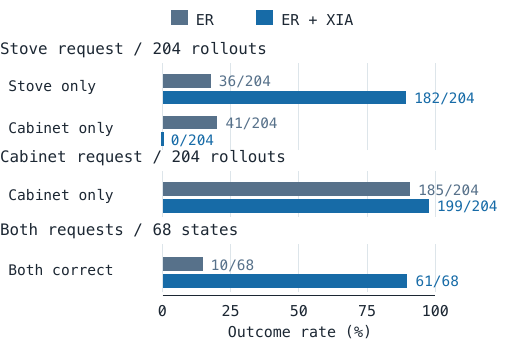}
\caption{\textbf{Goal selection from matched states.}
QwenVLA Goal after ten tasks, seeds 1337 and 2024. We run each
instruction three times from each of 68 states (204 rollouts per
instruction and method). A state counts as correct for both instructions
when the policy follows each one in at least two of its three rollouts.
Under the stove instruction, the remaining rollouts ended with no goal
satisfied (ER: 89; XIA: 22) or with another or multiple goals satisfied
(ER: 38; XIA: 0).}
\label{fig:supp-paired-instruction}
\end{figure}

We start from saved post-lift states from the stove and cabinet tasks.
From each state, the two instructions share the initial non-language
observation and matched sampling seeds; later observations follow
each rollout. At the ten-task endpoint, XIA improves correct selection
under the old stove instruction and reduces selection of the competing
cabinet goal. Correct selection under the cabinet instruction is also
higher (Fig.~\ref{fig:supp-paired-instruction}).

\FloatBarrier

\subsection{Query-Source and Baseline Controls}
\input{tables/query_source_controls}

\paragraph{Matched-parent queries.}
During the transition from the fourth to the fifth Goal task
(b3$\rightarrow$t4), the three arms share a teacher, 100 retained
frames per old task, and 10,000 updates within each seed. Each arm samples
four observation--instruction pairs per update from a pool of 400, with
four fixed noise/time tuples per pair.
The hybrid replaces the stored observation with a gripper-phase-matched
current observation but keeps the old action endpoint. It improves old-task
success over stored-old in two seeds (+27.5 and +21.0 percentage points)
and changes it by $-0.5$ points in the third. With current observations
fixed, swapping the endpoint changes old-task success by 0.5--1.5
points in these seeds.

\paragraph{Info-VLA adaptation.}
\label{sec:info-vla-spatial}
We adapt the released Info-VLA objective~\citep{zhao2026infovla} to
QwenVLA's trainable LoRA and flow head. A frozen boundary teacher supplies
visual replay targets. Language and action features from the main
minibatch enter the cross-modal losses. We use temperature 0.07,
detached loss-ratio normalization capped at 10, and an overall coefficient
of 0.1. Our adapted Info-VLA baseline uses the same frame-replay protocol as the other baselines.

\paragraph{SAM and DER++-style controls.}
SAM uses $\rho=0.075$ over trainable LoRA and action-head parameters.
Both gradient evaluations share the minibatch and restored stochastic
inputs. DER++-style adds 0.5 times the cached-velocity MSE on retained
frames, reusing the original noise and flow time.

\subsection{Cost of Anchoring}
ER and XIA retain the same expert frames. QwenVLA also caches current-task
anchors and their teacher targets for XIA (Table~\ref{tab:resource-accounting}).
Table~\ref{tab:measured-update-cost} counts input examples across forward
passes: SAM processes eight examples twice; XIA adds four anchor observations
to eight supervised examples, with four flow queries per anchor.
On an A800, the median update takes 0.75\,s for ER and 1.27\,s for XIA.
Building the initial XIA cache takes 319\,s.
\input{tables/resource_accounting}
\newpage
\input{tables/measured_update_cost}
\input{sec/S_franka}

%% file: sec/A_protocol.tex
\section{Implementation and Evaluation}
\label{sec:protocol-details}

Table~\ref{tab:training-settings} lists the LIBERO training settings.
Within each seed, ER and XIA share replay frames; XIA also matches
the previous policy on current observations paired with old instructions.

\input{tables/training_settings}

\subsection{Data, Replay, and Task Order}

We train on the no-noop LIBERO demonstrations with two 224-pixel views,
one external and one wrist-mounted. Each stored frame includes its observation, canonical
instruction, and expert action chunk. ER and XIA retain identical
100-frame samples per old task within each training seed; retrieval is
uniform over the retained buffer. The main minibatch is split equally
between current and replay examples once replay is available.
The first task uses only current examples. We keep the final minibatch
even when it is smaller. By the tenth task, the buffer holds 900 frames
from the nine previous tasks.

\paragraph{Task order.}
The dataset task IDs, in training order, are
\textbf{Goal}: 9, 7, 4, 2, 8, 5, 3, 6, 0, 1;
\textbf{Object}: 4, 2, 6, 3, 1, 8, 7, 5, 9, 0; and
\textbf{Spatial}: 4, 7, 8, 5, 1, 2, 0, 3, 6, 9.
In the result tables, t0--t9 index tasks in training order, and b$j$
denotes the checkpoint after task t$j$. All seeds use the orders listed above.
The full-sequence 100- and 1,000-frame buffers are independently sampled;
the matched-parent large-buffer Goal branch instead augments the original
100-frame buffer to 1,000.

\subsection{XIA Query Construction and Lifecycle}
\label{sec:xia-protocol}

\paragraph{Observation and instruction sampling.}
From the second task onward, we sample up to 1,000 frames uniformly
without replacement from the current demonstrations. For each frame, we
independently draw one old instruction uniformly and keep that pairing
fixed throughout the task.
On Goal, a fixed lookup maps all ten commands to the same kitchen scene,
so every previously learned command is eligible. Object and Spatial
also allow all previous commands.
Each update draws four distinct anchors uniformly from this pool.

\paragraph{Flow queries and targets.}
The current expert action defines the flow query; the frozen teacher
supplies the target response under the old instruction.
Noise is standard Gaussian, with the same chunk shape
as the normalized action. Each policy uses its native flow-time convention.
For QwenVLA, $z_t=(1-t)\epsilon+ta$ and
$t=(s-\min(B,s))/s$, where $B\sim\mathrm{Beta}(1.5,1)$ and $s=0.999$.
For $\pi_{0.5}$, $z_t=t\epsilon+(1-t)a$ and
$t=0.999B+0.001$. Teacher and student receive the same endpoint, noise,
time, observation, and old command. The auxiliary MSE is averaged over
sampled frames, noise/time tuples, action steps, and output coordinates.

\paragraph{Teacher targets.}
QwenVLA materializes each anchor image once and records four
noise/time/teacher-velocity tuples per frame with the boundary model in
evaluation mode. All four tuples are used when that frame is sampled.
Teacher velocities and noise are stored in half precision and converted
to float32 for loss evaluation. The student reuses the same image and
tuples; its training-mode dropout remains active.
For $\pi_{0.5}$ and OpenVLA-OFT, the frozen teacher and student process
the same image at each update. $\pi_{0.5}$ uses one fixed noise/time
tuple per anchor; OpenVLA-OFT matches normalized eight-step action
predictions with mean L1 loss. All three policies use $\alpha=1$.

\paragraph{Boundary lifecycle.}
After each task, ER updates the replay buffer. Before learning the next
task, XIA freezes the preceding policy as its teacher and builds a fresh
anchor pool. The teacher and query assignments stay fixed for that task,
as do the noise/time tuples for flow policies. At test time, only the
student acts.

\subsection{Controlled Comparisons and Outcomes}

\paragraph{Matched-parent controls.}
Local query-source controls share the starting policy, buffer, frozen
teacher and update count. Full-sequence text ablations share the current-frame
sampling rule. Before each subsequent task, EWC re-estimates the diagonal
Fisher from 100 two-frame minibatches drawn from the pooled replay buffer,
normalizes it to unit mean, and regularizes LoRA and action-head parameters
with $\lambda=0.001$.
Extra replay uses four current and eight old rows instead of four each.

\paragraph{Behavioral outcomes.}
\emph{Instruction takeover} occurs when a policy pursues a later-learned
goal under an earlier instruction. In Figure~2 of the main paper, each
off-diagonal entry counts rollouts that end with the column goal satisfied
under a different instruction. For example, under ER-100, 20 of 50
rollouts prompted with T2 end with the T9 goal satisfied.

\subsection{RoboCasa Implementation}
\label{sec:robocasa-implementation}

\begin{samepage}
We use RoboCasa v0.2~\citep{nasiriany2024robocasa} with the PandaOmron
embodiment. The five tasks, in the order listed in the main paper, use
52, 52, 53, 53, and 52 human demonstrations, respectively.
GR00T N1.5-3B takes two external RGB views, one wrist view, robot state,
and the task instruction. We freeze its vision-language backbone and
train the full action module. The policy predicts 16-step action
chunks and executes the first 10 controls before replanning.
\par
\end{samepage}

Both methods start from the same first-task checkpoint: 1,000 updates
at batch size 4 (100-step warmup to $3\!\times\!10^{-5}$), then 500
updates at batch size 8 and $10^{-5}$. Each later task uses 1,500 AdamW
updates at $10^{-5}$, with $(\beta_1,\beta_2)=(0.95,0.999)$ and global
gradient clipping at 1.0. The eight supervised examples per update
split evenly between current data and replay. The two arms use the
same supervised sample stream and retain 100 distinct expert replay
frames per prior task, with their native expert action chunks.

XIA pairs 1,000 distinct current observations with prior instructions,
balanced across old tasks, and uses four queries per update. The student
and its frozen previous-task teacher share the flow input
$z_t=(1-t)\epsilon+ta$ and time $t$, with current expert chunk $a$ and
Gaussian noise $\epsilon$. The loss is mean squared error between the
velocity predictions over valid action coordinates, with $\alpha=1$.
We report one paired training seed and 50 final rollouts per task and
method from matched initial states, with at most 480 controls per
rollout. For microwave deactivation, we reject states that already
satisfy the native success predicate before any policy action.

%% file: tables/training_settings.tex
\begin{table*}[!t]
\centering
\caption{\textbf{Full-sequence training settings on LIBERO.}
ER and XIA share the backbone configuration and expert-frame budget.
XIA adds the queries specified below.}
\label{tab:training-settings}
\small
\setlength{\tabcolsep}{5pt}
\renewcommand{\arraystretch}{1.15}
\begin{tabularx}{.94\textwidth}{@{}L{0.185\textwidth}*{3}{>{\raggedright\arraybackslash}X}@{}}
\toprule
Setting & QwenVLA & $\pi_{0.5}$ & OpenVLA-OFT\\
\midrule
Initialization & Qwen2.5-VL-3B-Instruct; new DiT action head & $\pi_{0.5}$ base & Generic OpenVLA-7B; OFT regression head\\
Adaptation & VLM LoRA rank 32, scale 16; train action head & VLM/expert LoRA ranks 16/32, scales 16/32; no EMA & VLM LoRA rank 32, scale 16; train action/proprioception heads\\
\addlinespace[2pt]
Updates per task & \textbf{10,000} & \textbf{7,500} & \textbf{10,000}\\
Main batch (current + old) & $4+4$ & $8+8$ & $4+4$\\
Action chunk length & 8 & 10 & 8\\
Learning rate & VLM $10^{-5}$; action head $10^{-4}$; other $2.5\!\times\!10^{-5}$ & $5\!\times\!10^{-5}$ & $5\!\times\!10^{-4}$\\
Schedule & 500-step warmup; task-wise cosine to $10^{-6}$ & 500-step initial warmup; constant thereafter & No warmup; $\times0.1$ at global step 100,000\\
\addlinespace[2pt]
AdamW $(\beta_1,\beta_2)$ & $(0.9,0.95)$ & $(0.9,0.95)$ & $(0.9,0.999)$\\
Weight decay; $\epsilon$ & $10^{-8}$; $10^{-8}$ & $10^{-10}$; $10^{-8}$ & $10^{-2}$; $10^{-8}$\\
Gradient norm clipping & 1 & 1 & None\\
Action normalization & q01/q99 for motion; binary gripper & Dataset quantile normalization & q01/q99 for motion; gripper unscaled\\
\midrule
\multicolumn{4}{l}{\textit{XIA: up to \textbf{1,000} current frames per task, \textbf{4} sampled frames per update, $\alpha=1$}}\\
Teacher response & Flow velocity & Flow velocity & Normalized action chunk\\
Response loss & Mean squared error & Mean squared error & Mean absolute error\\
Noise/time tuples per frame & 4, fixed & 1, fixed & Not applicable\\
Teacher evaluation & Cached at boundary & Online, shared augmented view & Online, shared augmented view\\
Image augmentation & None & Crop/resize, rotation, color jitter & Resized crop and color jitter\\
\bottomrule
\end{tabularx}
\end{table*}

%% file: tables/sequence_forgetting.tex
\begin{table*}[!tbp]
\centering
\caption{\textbf{QwenVLA acquisition, forgetting, and final success.}
Mean $\pm$ sample SD over seeds 42, 1337, and 2024.
Each matrix has 55 cells and 50 episodes per cell. Bold marks the better mean within each suite.}
\label{tab:sequence-forgetting}
\small
\setlength{\tabcolsep}{6pt}
\renewcommand{\arraystretch}{1.10}
\begin{tabular}{@{}ll@{\hspace{10pt}}cccc@{}}
\toprule
Suite & Method & Acquisition (\%)\enspace$\uparrow$ & NBT\enspace$\downarrow$ & Normalized NBT\enspace$\downarrow$ & Final success (\%)\enspace$\uparrow$\\
\midrule
\multirow{2}{*}{Goal} & ER & \meanstd{86.3}{2.3} & \meanstd{.446}{.063} & \meanstd{.522}{.065} & \meanstd{44.0}{3.6}\\
& ER + XIA & \meanstd{\textbf{89.2}}{0.4} & \meanstd{\textbf{.164}}{.060} & \meanstd{\textbf{.186}}{.060} & \meanstd{\textbf{71.4}}{8.8}\\
\addlinespace[3pt]
\multirow{2}{*}{Object} & ER & \meanstd{94.3}{2.5} & \meanstd{.329}{.016} & \meanstd{.353}{.026} & \meanstd{62.8}{3.0}\\
& ER + XIA & \meanstd{\textbf{94.5}}{0.6} & \meanstd{\textbf{.181}}{.026} & \meanstd{\textbf{.192}}{.023} & \meanstd{\textbf{74.7}}{4.5}\\
\addlinespace[3pt]
\multirow{2}{*}{Spatial} & ER & \meanstd{84.5}{3.2} & \meanstd{.368}{.038} & \meanstd{.456}{.041} & \meanstd{51.4}{2.4}\\
& ER + XIA & \meanstd{\textbf{89.8}}{2.4} & \meanstd{\textbf{.221}}{.005} & \meanstd{\textbf{.242}}{.010} & \meanstd{\textbf{68.5}}{2.7}\\
\bottomrule
\end{tabular}
\end{table*}

%% file: tables/query_source_controls.tex
\begin{table}[!tbp]
\centering
\caption{\textbf{Effect of observation source at a matched parent.}
Same-parent QwenVLA Goal update, b3$\rightarrow$t4, with historical
instructions in all arms. Entries count successful old-task (/200) and
new-task (/50) episodes, mean $\pm$ sample SD over three paired seeds.
Bold marks column maxima.}
\label{tab:query-source-controls}
\small
\setlength{\tabcolsep}{4pt}
\renewcommand{\arraystretch}{1.12}
\begin{tabular}{@{}llcc@{}}
\toprule
Query image & Flow endpoint &
\shortstack{Old successes\\(/200)\,$\uparrow$} &
\shortstack{New successes\\(/50)\,$\uparrow$}\\
\midrule
Stored old & Old action & \meanstd{33.3}{20.7} & \meanstd{44.0}{1.7}\\
Current & Current action & \meanstd{\textbf{67.3}}{24.0} & \meanstd{\textbf{45.0}}{2.0}\\
Current & Old action & \meanstd{65.3}{23.1} & \meanstd{44.7}{2.3}\\
\bottomrule
\end{tabular}
\end{table}

%% file: tables/resource_accounting.tex
\begin{table}[!htbp]
\centering
\caption{\textbf{Replay storage and training inputs for QwenVLA Goal.}
Frame counts use the nominal per-task budgets.
Storage counts the replay and anchor data in decimal MB. The last column
counts input examples per optimizer update.}
\label{tab:resource-accounting}
\small
\setlength{\tabcolsep}{3pt}
\begin{tabular}{@{}lrrrr@{}}
\toprule
Method & \shortstack{Old frames\\at last\\update} & \shortstack{Old\\payload\\(MB)} & \shortstack{Extra anchor/\\target payload\\(MB)} & \shortstack{Examples\\per update}\\
\midrule
ER-100 & 900 & 271.1 & 0 & 8 \\
ER + XIA & 900 & 271.1 & \textbf{301.2 + 0.91} & \textbf{12} \\
ER-1000 & 9,000 & 2710.5 & 0 & 8 \\
\bottomrule
\end{tabular}
\end{table}

%% file: tables/measured_update_cost.tex
\begin{table}[!t]
\centering
\caption{\textbf{QwenVLA success and update cost on LIBERO-Goal.}
Final SR is the mean $\pm$ SD across training seeds after all ten tasks.
Timing is measured separately on an A800: medians over 51 update intervals
(52 for DER++-style and adapted Info-VLA), relative to ER at 0.75\,s/update.
Best values are bold.}
\label{tab:measured-update-cost}
\footnotesize
\setlength{\tabcolsep}{2pt}
\begin{tabular}{@{}lrrrrr@{}}
\toprule
Method & \shortstack{Final SR\\(\%) $\uparrow$} & \shortstack{Time\\(s) $\downarrow$} & \shortstack{Time/\\ER $\downarrow$} & \shortstack{Peak GPU\\(MiB) $\downarrow$} & \shortstack{Examples\\per update $\downarrow$}\\
\midrule
\rowcolor{ResultGray}ER & \meanstd{44.0}{3.6} & \textbf{0.75} & \textbf{1.00} & \textbf{35,960} & \textbf{8} \\
ER, extra replay & \meanstd{42.0}{4.0} & 0.97 & 1.29 & 47,704 & 12 \\
ER + EWC & \meanstd{61.7}{5.3} & 0.98 & 1.31 & 38,560 & \textbf{8} \\
ER + SAM & \meanstd{50.5}{5.3} & 1.68 & 2.24 & 41,202 & 16 \\
DER++-style & \meanstd{44.2}{8.8} & 1.18 & 1.58 & 46,166 & 12 \\
Info-VLA (adapted) & \meanstd{44.9}{8.9} & 1.07 & 1.43 & 44,572 & 16 \\
\rowcolor{ResultBlueLight}ER + XIA & \meanstd{\textbf{71.4}}{8.8} & 1.27 & 1.69 & 46,250 & 12 \\
\bottomrule
\end{tabular}
\end{table}

%% file: sec/S_franka.tex
\begin{figure*}[!t]
\centering
\begin{minipage}[c]{.43\textwidth}
  \centering
  \includegraphics[width=.96\linewidth,trim=200 65 0 110,clip]{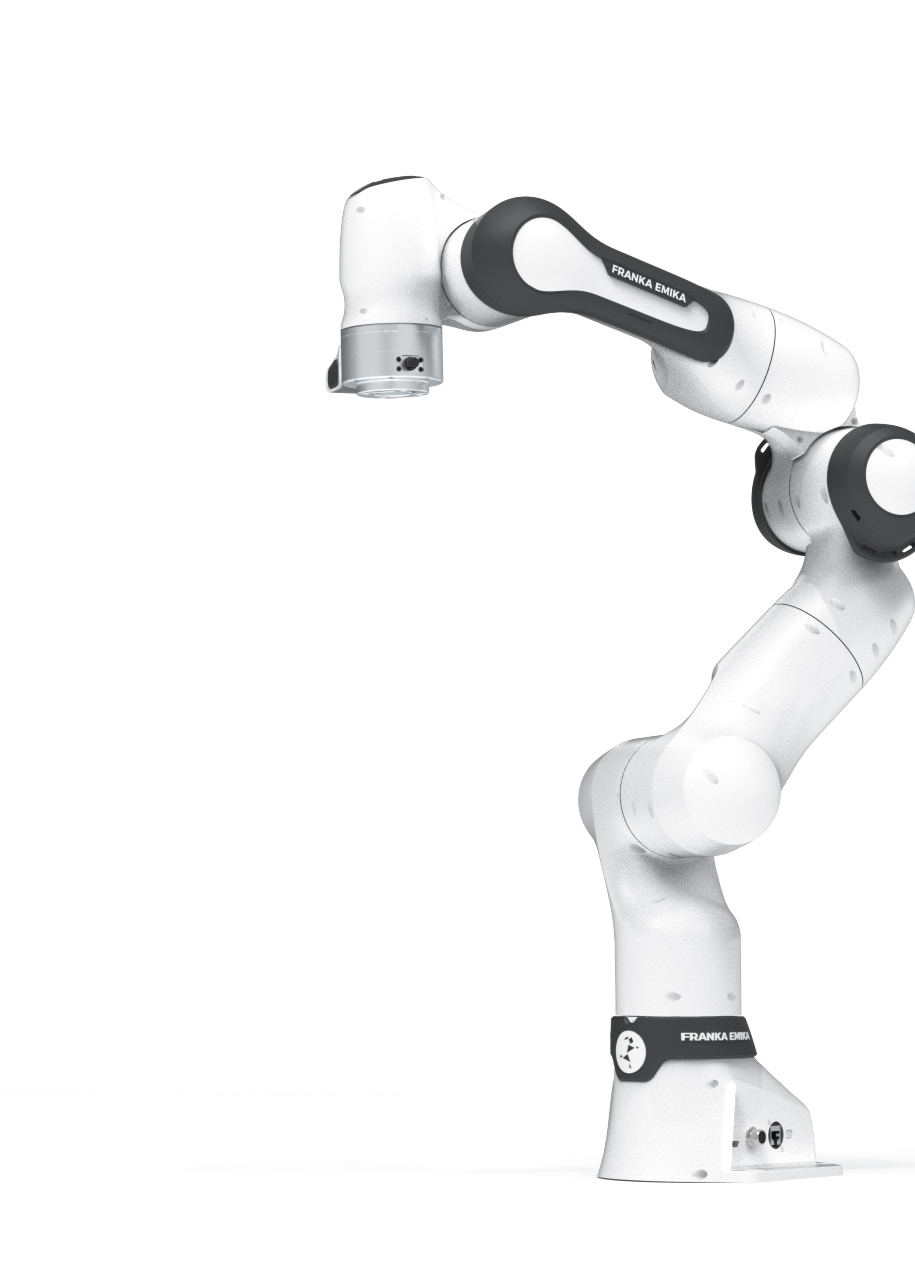}\par
  \small\textbf{(a) FR3 arm (manufacturer reference)}
\end{minipage}\hfill
\begin{minipage}[c]{.53\textwidth}
  \centering
  \includegraphics[width=.87\linewidth]{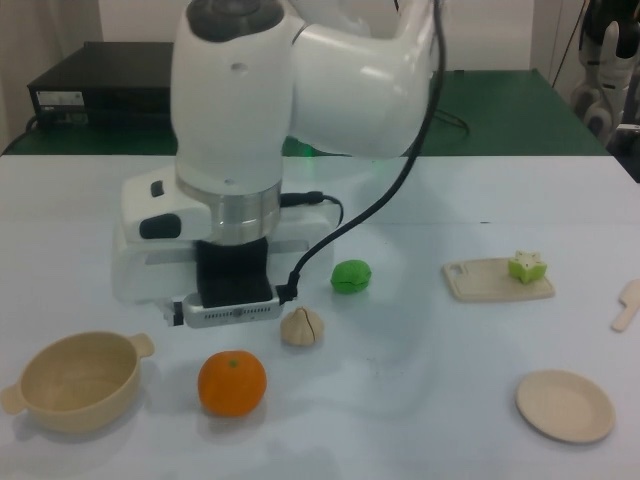}\par
  \small\textbf{(b) External RGB view}\par\medskip
  \includegraphics[width=.87\linewidth]{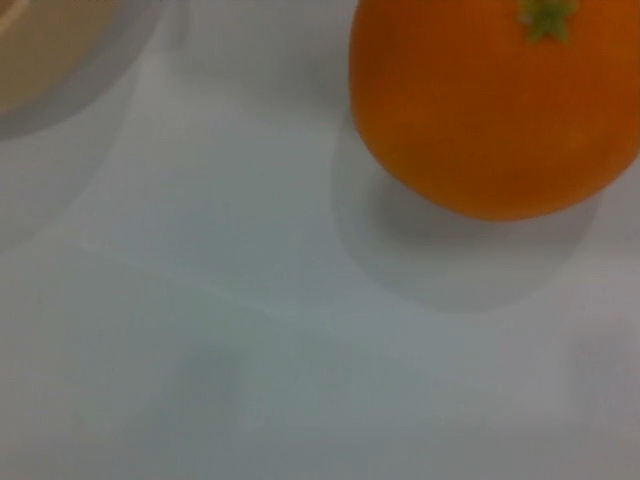}\par
  \small\textbf{(c) Wrist RGB view}
\end{minipage}
\caption{\textbf{Franka Research 3 platform and policy observations.}
(a) Arm shown in the Franka Robotics \emph{Franka Research 3}
datasheet (R02212, rev.~1.2).
(b,c) Same-timestamp frames from the orange-to-board demonstration in
our cell. The external view shows the tabletop and destinations; the
wrist view shows the workspace near the gripper.}
\label{fig:fr3-platform}
\end{figure*}

\newcommand{\FrankaFrame}[2]{%
  \begin{minipage}[t]{.238\textwidth}
    \centering#1\par
    \footnotesize #2\,s
  \end{minipage}}
\newcommand{\FrankaRow}[2]{%
  \noindent\textbf{#1}\par\smallskip
  \noindent#2\par}

\begin{figure*}[!t]
\FrankaRow{T1. Place the orange on the cutting board}{%
  \FrankaFrame{\includegraphics[width=\linewidth]{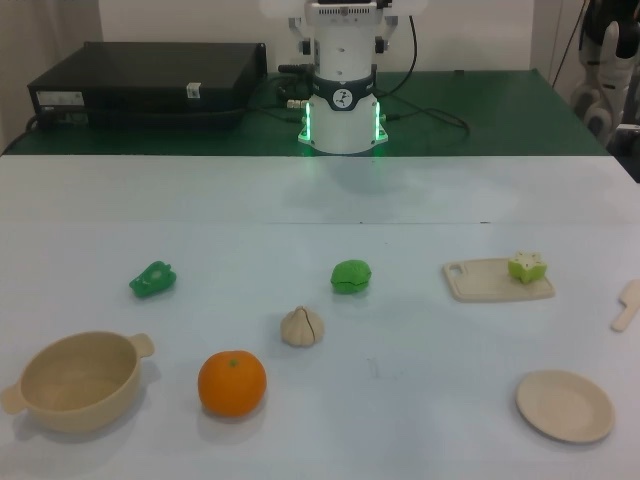}}{0.0}\hfill
  \FrankaFrame{\includegraphics[width=\linewidth]{figures/fr3_supplement_20261118/demo_board/frame_02.jpg}}{5.6}\hfill
  \FrankaFrame{\includegraphics[width=\linewidth]{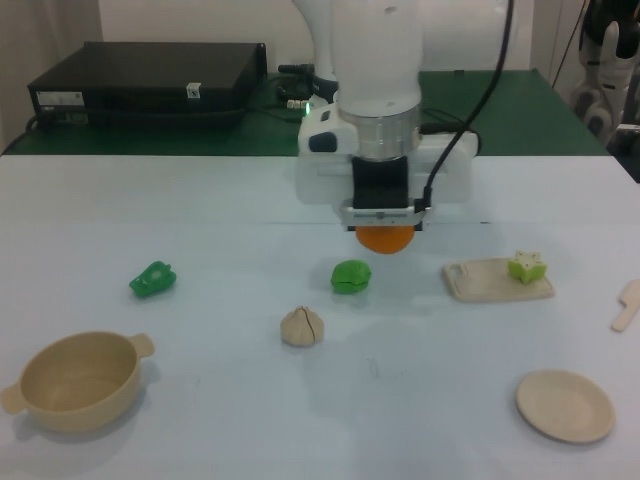}}{11.2}\hfill
  \FrankaFrame{\includegraphics[width=\linewidth]{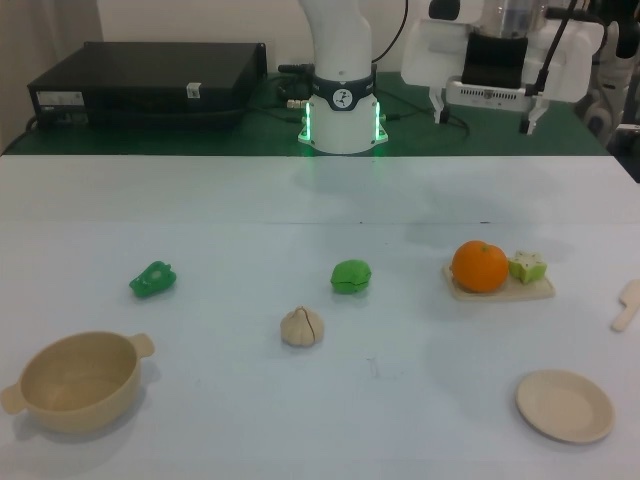}}{16.9}}
\medskip
\FrankaRow{T2. Transfer the sliced starfruit to the plate}{%
  \FrankaFrame{\includegraphics[width=\linewidth]{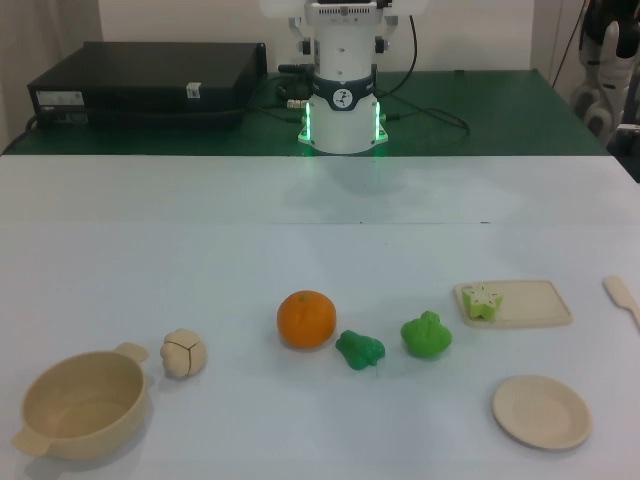}}{0.0}\hfill
  \FrankaFrame{\includegraphics[width=\linewidth]{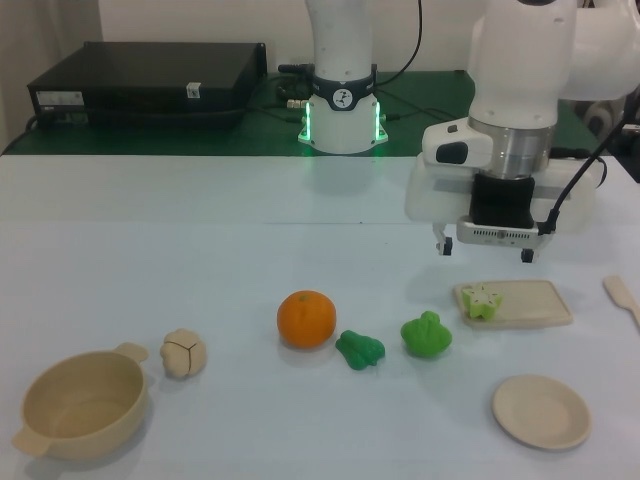}}{6.2}\hfill
  \FrankaFrame{\includegraphics[width=\linewidth]{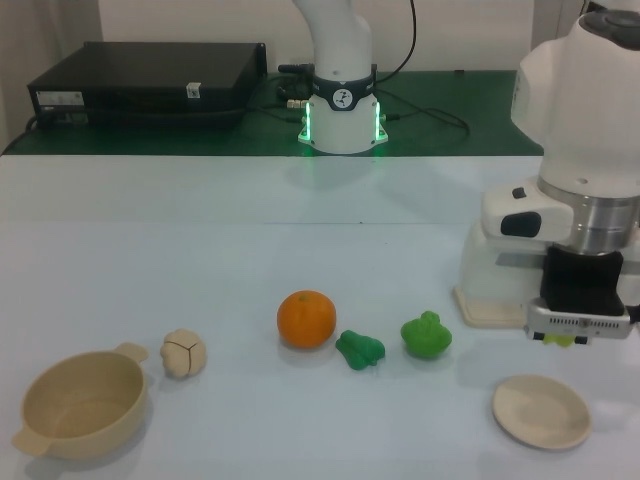}}{12.3}\hfill
  \FrankaFrame{\includegraphics[width=\linewidth]{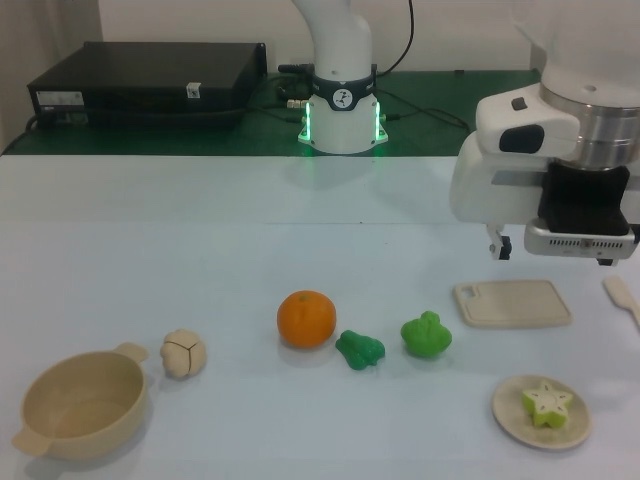}}{18.5}}
\medskip
\FrankaRow{T3. Place the orange in the bowl}{%
  \FrankaFrame{\includegraphics[width=\linewidth]{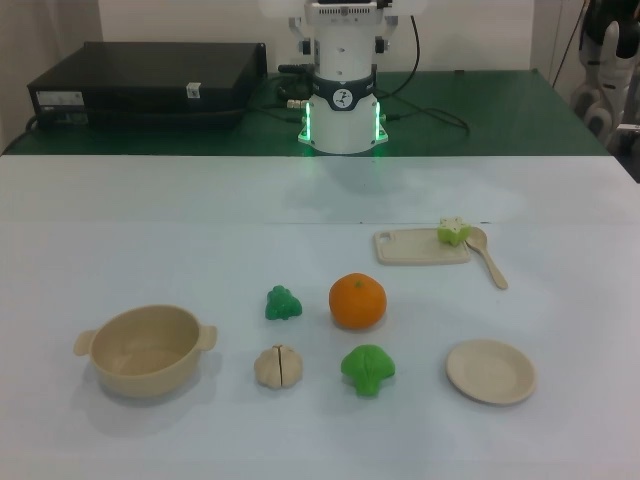}}{0.0}\hfill
  \FrankaFrame{\includegraphics[width=\linewidth]{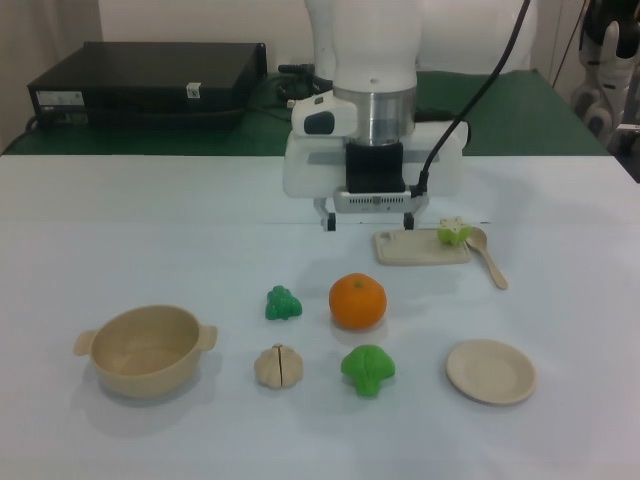}}{3.8}\hfill
  \FrankaFrame{\includegraphics[width=\linewidth]{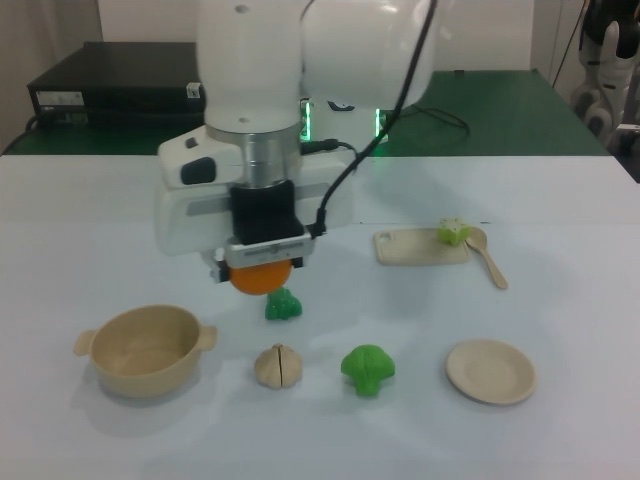}}{7.6}\hfill
  \FrankaFrame{\includegraphics[width=\linewidth]{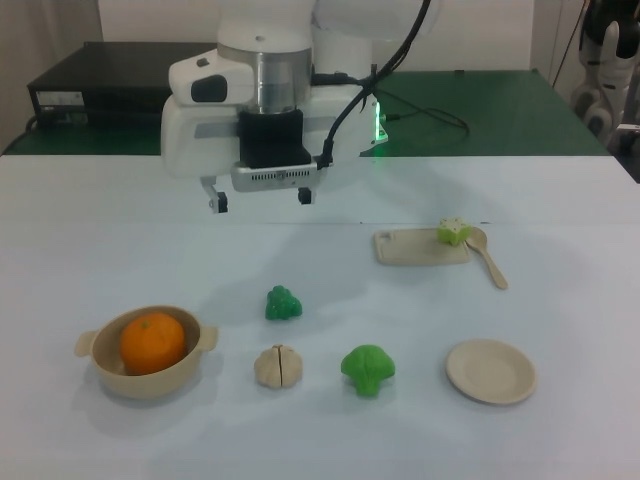}}{11.4}}
\caption{\textbf{Real-robot task demonstrations.}
Each row shows four chronological external-camera frames from one
demonstration. Times are relative to the start of the corresponding
video. T1 and T3 use the same object but specify different destinations.}
\label{fig:fr3-demonstrations}
\end{figure*}

\section{Real-Robot Platform and Rollouts}
\label{sec:franka-supplement}

\paragraph{Hardware and observations.}
The FR3 policy takes external and wrist RGB views (Fig.~\ref{fig:fr3-platform}),
padded and resized to $224\times224$, and joint and gripper positions.
It commands seven joint velocities and the parallel-jaw gripper.

\par\medskip
\noindent\begin{minipage}{\columnwidth}
\centering
\captionof{table}{\textbf{Franka training and evaluation protocol.}
The three tasks use 184 expert demonstrations. Both methods start from
DROID-pretrained $\pi_{0.5}$ and share the first-task checkpoint.}
\label{tab:fr3-protocol}
\small
\setlength{\tabcolsep}{2.5pt}
\begin{tabularx}{\linewidth}{@{}L{.265\linewidth}YYY@{}}
\toprule
\shortstack[l]{\textbf{Task}\\\footnotesize\phantom{X}}
& \shortstack{\textbf{T1}\\\footnotesize\textcolor{ResultGrayDark}{Orange to board}}
& \shortstack{\textbf{T2}\\\footnotesize\textcolor{ResultGrayDark}{Starfruit to plate}}
& \shortstack{\textbf{T3}\\\footnotesize\textcolor{ResultGrayDark}{Orange to bowl}} \\
\cmidrule(lr){2-2}\cmidrule(lr){3-3}\cmidrule(l){4-4}
Expert demos & \textbf{47} & \textbf{71} & \textbf{66} \\
\end{tabularx}\par\nointerlineskip
\begin{tabularx}{\linewidth}{@{}L{.265\linewidth}>{\raggedright\arraybackslash}X@{}}
\midrule
Updates & 10,000 per task \\
Minibatch & 32 (16 current + 16 replay after T1) \\
Replay memory & 100 frames per previous task \\
Learning rate & 1k-step warmup to $2.5\!\times\!10^{-5}$;
cosine decay to $2.5\!\times\!10^{-6}$ at 30k \\
\midrule
Control rate & 15\,Hz \\
Replanning & After 8 of 16 predicted actions \\
Evaluation & 20 trials per seen task per stage \\
\bottomrule
\end{tabularx}
\end{minipage}
\par\medskip

On each later task, XIA pairs 1,000 distinct current frames with old
instructions sampled uniformly, fixing one noise/time tuple per frame.
Each update samples four distinct anchors; the frozen teacher supplies
their velocity targets online ($\alpha=1$).

\begin{samepage}
\paragraph{Evaluation protocol.}
Before each trial, the operator resets the tabletop objects and the arm
returns to its saved joint pose. A trial succeeds when the robot places
the requested object at the instructed destination. After T3, XIA succeeds
in 55/60 trials versus 36/60 for ER. The largest difference is on T1
(19/20 vs.\ 4/20); both methods succeed in all 20 T3 trials.
\par
\end{samepage}

\par\medskip
\noindent\begin{minipage}{\columnwidth}
\paragraph{Rollout gallery.}
Figures~\ref{fig:fr3-rollouts} and~\ref{fig:fr3-rollouts-final} show one
rollout per evaluated instruction and arm. The first checkpoint is shared:
the gallery has one rollout after T1, four after T2, and six after T3.
The matrix below indexes the filmstrips by evaluated instruction. Cells
with two IDs list ER first and ER+XIA second.

\nopagebreak[4]
{\centering\small
\setlength{\tabcolsep}{5pt}
\begin{tabular}{@{}lccc@{}}
\toprule
After training & T1: board & T2: plate & T3: bowl \\
\midrule
T1 (10k) & \textbf{01} (shared) & -- & -- \\
T2 (20k) & \textbf{02 / 03} & \textbf{04 / 05} & -- \\
T3 (30k) & \textbf{06 / 07} & \textbf{08 / 09} & \textbf{10 / 11} \\
\bottomrule
\end{tabular}\par}

\smallskip
Each strip shows one trial, labeled by its outcome.
\par
\end{minipage}
\par\medskip

\input{sec/S_franka_matrix}

%% file: sec/S_franka_matrix.tex
\begingroup
\ifdefined\overleafshowgallery
  \setkeys{Gin}{draft=false}
\fi
\begin{figure*}[p]
\centering
\includegraphics[width=\textwidth]{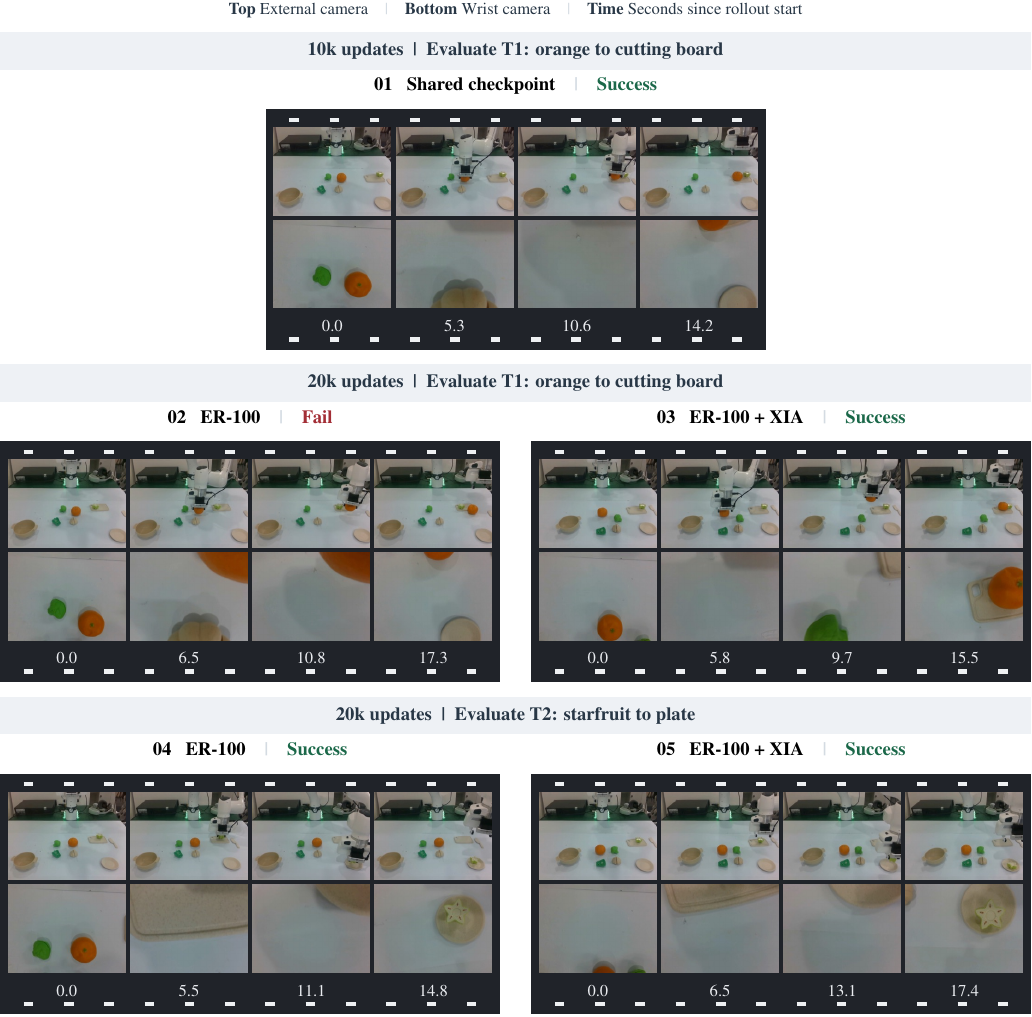}
\caption{\textbf{Real-robot rollouts after the first and second tasks.}
Each numbered strip shows four chronological snapshots of one episode.
The two camera views in each column share a timestamp. The 10k
checkpoint is common to both methods. Green Success and red Fail
refer to completion of the displayed instruction in that episode.}
\label{fig:fr3-rollouts}
\end{figure*}

\providecommand{\lastrolloutpanel}[1]{\begin{figure*}[p]#1\end{figure*}}
\lastrolloutpanel{%
\centering
\includegraphics[width=\textwidth]{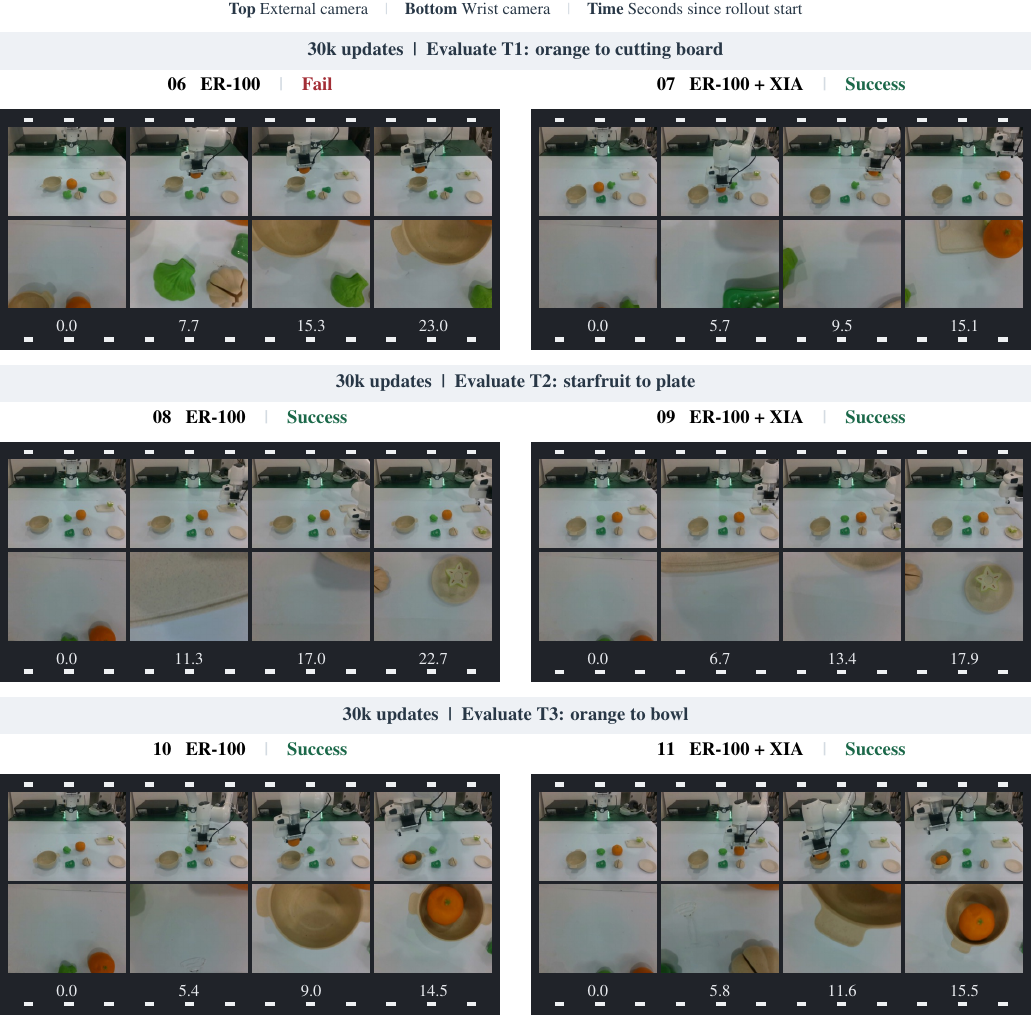}
\caption{\textbf{Real-robot rollouts after the third task.}
Each row shows ER and XIA performing the same task.
For visual clarity, all gallery frames are processed with the same mild exposure adjustment.}
\label{fig:fr3-rollouts-final}
}
\endgroup